\documentclass[10pt,twocolumn,letterpaper]{article}

\usepackage[accsupp]{axessibility}
\usepackage{iccv}
\usepackage{times}
\usepackage{epsfig}
\usepackage{graphicx}
\usepackage{amsmath}
\usepackage{amssymb}
\usepackage{booktabs}
\usepackage[table,xcdraw]{xcolor}
\usepackage{rotating}
\usepackage{subcaption}
\usepackage{multirow}
\usepackage{hyperref}
\hypersetup{
    colorlinks=true,
    linkcolor=red,
    filecolor=magenta,      
    urlcolor=cyan,
    pdftitle={AdVerb},
    pdfpagemode=FullScreen,
    }
\usepackage{pifont}% http://ctan.org/pkg/pifont
\usepackage[accsupp]{axessibility}
\newcommand{\xmark}{\ding{55}}%

\newcommand\modelname{AdVerb}
\newcommand\novellossone{Spectrogram Prediction Loss}
\newcommand\novellosstwo{Acoustic Token Matching Loss}

\iccvfinalcopy % *** Uncomment this line for the final submission

\ificcvfinal\fi

\begin{document}

%%%%%%%%% TITLE
\title{\modelname: Visually Guided Audio Dereverberation}

\author{Sanjoy Chowdhury$^{1}\thanks{Equal contribution.}$ $\quad$ Sreyan Ghosh$^{1*}$ $\quad$ Subhrajyoti Dasgupta$^2$ \\ Anton Ratnarajah$^1$ $\quad$ Utkarsh Tyagi$^1$ $\quad$ Dinesh Manocha$^1$ \vspace{2mm} \\ 
$^{1}$University of Maryland, College Park $\quad$ $^{2}$Mila and Université de Montréal \vspace{2mm}\\ 
% Institution1 address\
\tt\small \{sanjoyc,sreyang,jeran,utkarsht,dmanocha\}@umd.edu $\quad$ subhrajyoti.dasgupta@umontreal.ca \\
\tt\small Project page -\ \url{https://gamma.umd.edu/researchdirections/speech/adverb}}

% \author{Sanjoy Chowdhury$^*$\\
% University of Maryland, College Park\\
% {\tt\small sanjoyc@umd.edu}
% \and
% Sreyan Ghosh$^*$\\
% University of Maryland, College Park\\
% % First line of institution2 address\\
% {\tt\small sreyang@umd.edu}
% \and
% Subhrajyoti Dasgupta\\
%  Mila and Université de Montréal\\
% % First line of institution2 address\\
% {\tt\small subhrajyoti.dasgupta@umontreal.ca}
% \and
% Anton Ratnarajah\\
% University of Maryland, College Park\\
% % First line of institution2 address\\
% {\tt\small jeran@umd.edu}
% \and
% Utkarsh Tyagi \\
%  MIDAS Labs, IIIT Delhi, India\\
% % First line of institution2 address\\
% {\tt\small utkarsh4430@gmail.com}
% \and
% Dinesh Manocha\\
% University of Maryland, College Park\\
% % First line of institution2 address\\
% {\tt\small dmanocha@umd.edu}
% }

\maketitle


%%%%%%%%% TITLE
% \title{\modelname: Visually Guided Audio Dereverberation}


% \and

% \maketitle

%%%%%%%%% ABSTRACT
\begin{abstract}

We present \modelname, a novel audio-visual dereverberation framework that uses visual cues in addition to the reverberant sound to estimate clean audio. Although audio-only dereverberation is a well-studied problem, our approach incorporates the complementary visual modality to perform audio dereverberation.  Given an image of the environment where the reverberated sound signal has been recorded, \modelname~ employs a novel geometry-aware cross-modal transformer architecture that captures scene geometry and audio-visual cross-modal relationship to generate a complex ideal ratio mask, which, when applied to the reverberant audio predicts the clean sound. The effectiveness of our method is demonstrated through extensive quantitative and qualitative evaluations. Our approach significantly outperforms traditional audio-only and audio-visual baselines on three downstream tasks: speech enhancement, speech recognition, and speaker verification, with relative improvements in the range of 18\% - 82\% on the LibriSpeech test-clean set. We also achieve highly satisfactory RT60 error scores on the AVSpeech dataset. 
\end{abstract}

% \blfootnote{$^*$These authors contributed equally to this work.}

% \let\thefootnote\relax\footnote{$^\ast$Equal contribution.}

% Real-world audio often gets corrupted by unwanted reverberation due to the persistence of sound after it has been stopped owing to multiple reflections from surfaces such as furniture, people, etc. These reflections build up with each reflection and decay gradually as they are absorbed by the surfaces of objects in the enclosed space and their removal is highly desirable. Although dereverberation of audio signals is a well-studied problem in the sound and speech processing domain, we notice the problem has not been investigated aptly under the audio-visual setting. To this end, we present \modelname~ to perform visually guided sound dereverberation. Our novel self-supervised framework extracts additional room cues in the form of horizontal depth along with room height to obtain a robust omnidirectional geometry-aware representation of an indoor scene in both horizontal and vertical directions. This coupled with two novel loss functions \novellossone~ and \novellosstwo~ achieves benchmark results on downstream practical tasks. We deploy a novel Transformer architecture consisting of \textit{(Shifted) window Blocks} and \textit{Global Blocks} to combine the local and global geometry relations. Further, we develop a novel relative position embedding of Transformer to enhance the spatial representation ability for the panoramic images.

% \vspace{-2mm}
\section{Introduction}

\begin{figure*}[t]
    \centering
    \includegraphics[width=\textwidth]{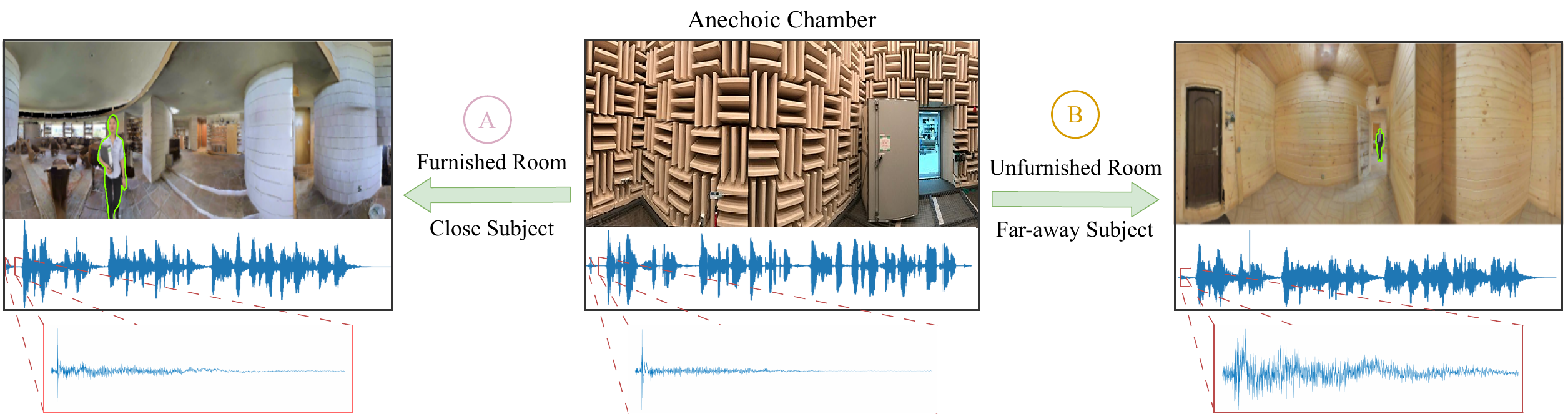}
    \vspace{-6mm}
    \caption{Reverberation is a function of the speaker's relative position and the surrounding environment. The visual signals present critical details that determine the nature of the distortion. E.g, \textcircled{\raisebox{-0.9pt}{A}} in a relatively small furnished room when the speaker is nearby, reverb is less evident, whereas for \textcircled{\raisebox{-0.9pt}{B}} in a large hallway (especially when the speaker is far away) the reverb effect is very strong. The audio waveform illustrates the nature of reverberation, with the magnified section clearly depicting a stronger reverberation effect in case \textcircled{\raisebox{-0.9pt}{B}} over \textcircled{\raisebox{-0.9pt}{A}}.}
    \label{fig:my_label}
    \vspace{-5mm}
\end{figure*}

Reverberation occurs when an audio signal reflects from multiple surfaces and objects in the environment to alter the dry sound thereby degrading its quality. Far-field speech recorded at a considerable distance from the speaker is significantly degraded by the strong reverberation effects caused by the environment. The amount of reverberation is highly correlated to the geometry of the surroundings and the materials present in the vicinity~\cite{chen2020soundspaces,chensoundspaces}. For instance, the auditory experience changes drastically when listening to a pleasant symphony in a large empty hallway vs. a relatively small furnished living room (Fig. \ref{fig:my_label}). Recent studies have shown that the reverberation effects can be estimated from a single image of the environment with reasonable accuracy \cite{singh2021image2reverb,majumder2022fewshot, Kon2019EstimationOL}. Removal of reverberation in recorded speech signals is highly desirable and would help improve the performance of several other auxiliary downstream tasks like automatic speech recognition (ASR), speaker verification (SV), source separation (SP), speech enhancement (SE), etc., which are widely used in several day-to-day tools.

% In this work, we build on these key observations and discuss this in Section \ref{section: methodology}.

% Inspired by the previous observations, we propose a novel audio-visual speech enhancement approach that captures the geometry and material information from the visual cues and helps our overall architecture to remove reverberation from the captured audio.      

% In real-world scenarios, audio often gets corrupted by unwanted reverberation owing to multiple reflections from surfaces such as furniture, people, etc. These reflections build up with each object and decay gradually as they are absorbed by the surfaces of objects in the enclosed space. Their removal is highly desirable for both better human perception of audio and intelligent systems that take audio as input \cite{}.

\begin{figure}
    \centering
    \includegraphics[width=8cm]{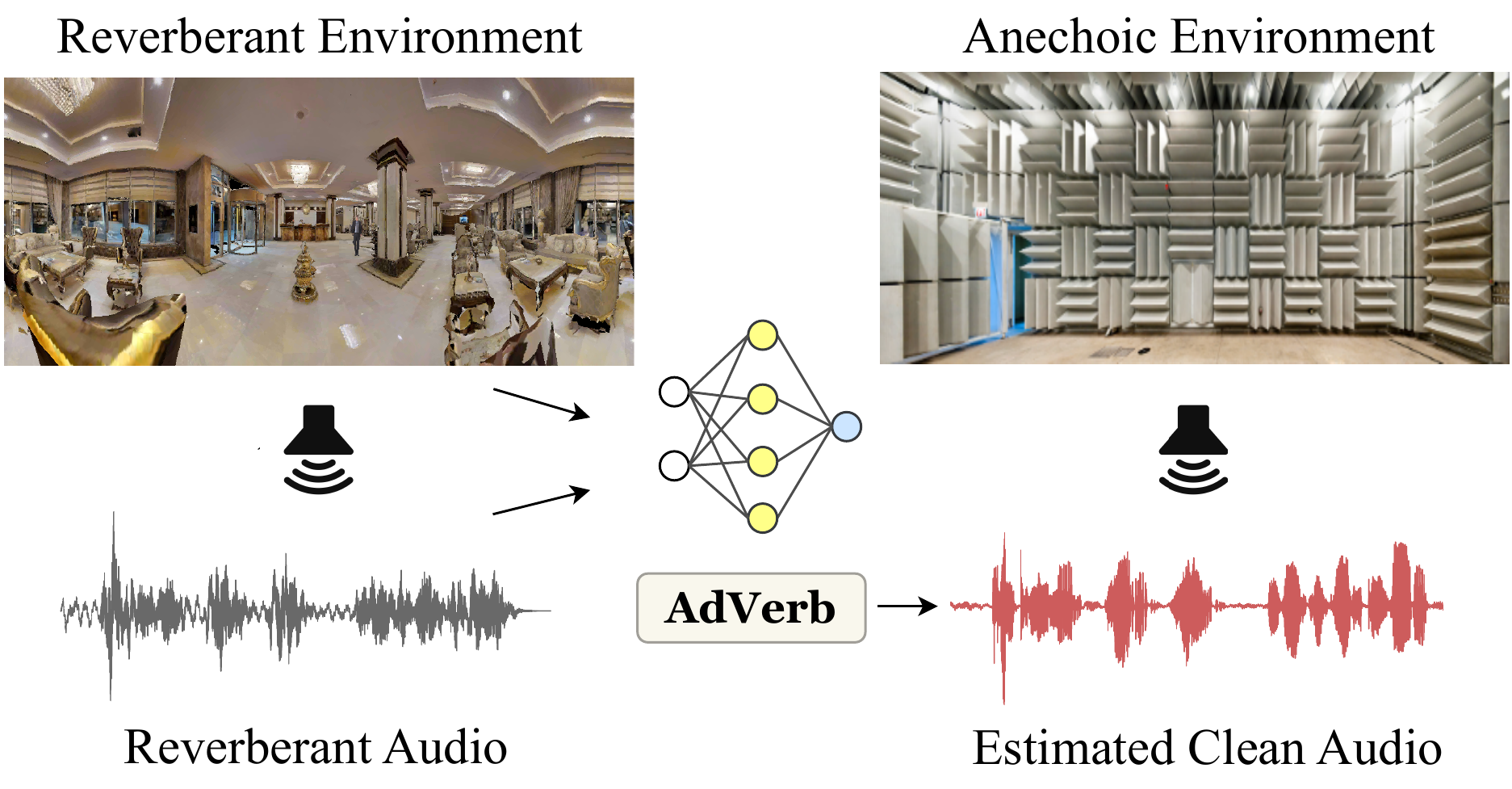}
    \vspace{-2mm}
    \caption{We present AdVerb, a novel audio-visual dereverberation framework that leverages visual cues of the environment to estimate  clean audio from reverberant audio. E.g, given a reverberant sound produced in a large hall, our model attempts to remove the reverb effect to predict the anechoic or clean audio.}
    \label{fig:title_diagram}
    \vspace{-5mm}
\end{figure}

% Dereveberation is a well-studied problem in the speech and signal processing community, with various audio-only systems achieving remarkable results \cite{naylor2010speech, kinoshita2016summary, zhao2018two, wu2016reverberation, wu2017end}. 
% Despite the fact that vision provides complementary cues for audio dereverberation 

% - reverberation highly depends on room geometry and surrounding material's properties and in most cases the reverberant audio is naturally companied with the visual stream (for e.g., video conferencing, augmented
% reality, etc.) we would like to emphasize that learning audio-visual dereverberation of far-field speech is an understudied problem and is challenging to tackle. 

Audio-only dereverberation is a well-studied problem with various systems achieving encouraging results \cite{naylor2010speech, kinoshita2016summary, zhao2018two, wu2016reverberation, wu2017end}. In contrast, using the visual stream as an additional cue to solve this task is a particularly understudied problem. We attribute the lack of research in this space to the scarcity of datasets. Most open-source datasets, both real and synthetic, contain only room impulse responses (RIRs) with no information about their source of origin \cite{szoke2019building,traer2016statistics}. Note that obtaining such RIRs can be challenging as doing so requires access to the physical environment, thereby limiting their applicability. However, in real-world settings, reverberant audio is naturally accompanied by a visual stream; video conferencing, augmented reality (AR), and web video indexing are some examples.

Recently, audio-visual speech enhancement methods~\cite{audio-visual-enhance1,audio-visual-enhance2,chen2021learning,audio-visual-enhance3} have shown significant improvements over the audio-only speech enhancement approaches. These tasks benefit from the presence of sound-producing objects in the visual scene, which allows the model to effectively utilize these strong stimuli for accomplishing the task. Many of these approaches track the lip movements of the speaker to separate the noise from the voice components in degraded speech which builds on the assumption that a speaker is always close to and facing the camera. These assumptions might not always hold in our case as the scope of the problem under consideration (mid/far-field) makes it difficult to obtain such cues. Thus, in a real-world setting, the available cue for audio-visual dereverberation is a panoramic view of the environment with or without a speaker in the field of view. Effectively utilizing visual cues in order to perform audio-visual dereverberation would require the model to understand the room's implicit geometric and material properties, which poses its own challenges.
\vspace{1mm}

{\noindent \textbf{Our Contributions:}} We propose \modelname, comprising a modified conformer block \cite{gulati2020conformer} with specially designed positional encoding to learn audio-visual dereverberation. The network takes corrupted audio and the corresponding visual image\footnote{We use panoramic images to train; inference can be done on both panoramic and non-panoramic images.} of the surrounding environment (from where the RIR is obtained) as input to perform this task (Fig. \ref{fig:title_diagram}). Our approach employs a \textit{novel geometry-aware module} with cross-modal attention between the  audio and visual modalities to generate a \textit{complex ideal ratio mask}, which is applied to the reverberant spectrogram to obtain the estimated clean spectrogram. This conformer block consists of a \textit{modified (Shifted) Window Block} \cite{liu2021swin} and \textit{Panoptic Blocks} to combine local and global geometry relations. We discuss key motivations behind our approach in Section \ref{section: methodology}. To learn audio-visual dereverberation, AdVerb solves two objectives, \textit{\novellossone~}and \textit{\novellosstwo}, which makes the output audio retain phonetic and prosodic properties. To summarize, our main contributions are as follows:

\textbf{(1)} We propose \modelname, \textit{a novel cross-modal framework} for dereverberating audio by exploiting complementary low-level visual cues  and specially designed relative position embedding.

\textbf{(2)} To this end, \modelname~employs \textit{a novel geometry-aware conformer network} to capture 3D spatial semantic information to equip the network with salient vision cues through (Shifted) Window Blocks and Panoptic Blocks.

\textbf{(3)} Our architecture involves the prediction of \textit{complex ideal ratio mask} and simultaneous optimization of two objective functions to estimate the dereverbed speech.

\textbf{(4)} On objective evaluation our approach significantly outperforms traditional audio-only and audio-visual \cite{chen2021learning} baselines with a relative improvement in the range 18\% - 82\% on three downstream tasks: speech enhancement, speech recognition, and speaker verification, when evaluated on the LibriSpeech test-clean set on all difficulty levels. It also achieves highly satisfactory RT60 error scores on the AVSpeech dataset.

\textbf{(5)} User study analysis reveals our method outperforms prior approaches on perceptual audio quality assessment.

\section{Related Works}

{\noindent \textbf{Audio Dereverberation:}} In communication and speech processing applications, reverberation can reduce intelligibility and weaken a dry audio signal \cite{naylor2010speech, kinoshita2016summary, zhao2018two, wu2016reverberation, wu2017end}. Lately, there has been a paradigm shift from using the traditional signal processing-based methods to neural networks and, subsequently, deep learning-based methods for dereverberation. Kinoshita \textit{et al.} \cite{kinoshita2017neural} presents a deep neural network to estimate the power spectrum of the target signal for weighted prediction error. Extending this, Wang \textit{et al.} \cite{wang2020deep} deploy a CNN-based model to separate the real and imaginary parts of clean speech. Typically, there are two prominent ways of training such models: through supervised learning \cite{xiao2016speech, luo2018real} or through adversarial networks (GANs) \cite{su2020hifi,su2021hifi}. Audio reverberation in nature is heavily influenced by room acoustics \cite{liu2020sound}. We find studies in the literature that try to capture room-specific information for finer modeling of acoustic environments \cite{su2020acoustic, giri2015improving}. Another line of work \cite{tan2020audio, li2022audio} attempts to extract visual features of target lip movements. Work from Chen \textit{et al.}~\cite{chen2021learning} is most similar in spirit to our proposed approach. These studies motivate us to pursue audio-visual dereverberation by leveraging room-aware geometric cues. Our framework exploits panoramic image features and is applicable even for out-of-view speaker cases. 
\vspace{2mm}

{\noindent \textbf{Room Impulse Response and Geometry Awareness:}} For a given environment, the amount of reverberation in the speech signal is mathematically described using a function known as room impulse response (RIR). RIR generators are used to simulate large-scale speech training data~\cite{ratnarajah21_interspeech,fast-rir}. While \cite{holters2009impulse, stan2002comparison,butreverb} engage dedicated in-room amenities to estimate this function, another line of research \cite{allen1979image, murphy2007acoustic, chen2020soundspaces,gwa,mesh2ir} choose to produce RIRs synthetically. These works \cite{kon2019estimation, singh2021image2reverb} estimate RIRs from an RGB and depth image. One downside of these approaches is that they require access to paired image and impulse response data. In contrast, some prior methods \cite{jeub2010we, jeub2009binaural, murphy2010openair} for generating RIR  operate by using images taken at arbitrary distances from the point of audio capture.     

Video streams, by nature, capture the natural association between visual and audio modalities. Wang \textit{et al.} \cite{wang2021led2} propose a geometry-aware approach for room layout estimation by horizon depth, which is only effective in the horizontal direction. Hu \textit{et al.} \cite{hu2019revisiting} and Eder \textit{et al.} \cite{eder2019pano} introduce gradient of depth and plane aware loss, respectively for improved depth estimation of panoramic images. These works inspire us to leverage room geometry to model this problem.
\vspace{0.3mm}

{\noindent \textbf{Audio-Visual Learning:}} Cross-modal learning powered by large-scale video datasets has been pushing boundaries in applications like audio-visual sound separation \cite{zhao2018sound, zhao2019sound, gao2019co, gan2020music, xu2019recursive}, audio-visual speech enhancement \cite{afouras2018conversation, afouras2019my, hegde2021visual, yang2022audio}, active speaker detection \cite{afouras2020self, alcazar2020active, tao2021someone, roth2020ava}, talking head generation \cite{wang2021audio2head, chen2020talking, prajwal2020lip}, embodied AI for audio-visual navigation \cite{chen2021semantic, chen2020learning, majumder2022active, yu2022sound}, etc. In addition, many recent works have utilized paired audio-visual data for representation learning. Owens \textit{et al.} \cite{owens2016visually} learned visual representations for materials from impact sounds. Another line of work learns features, scene structure, and geometric properties \cite{chen2021structure, owens2016ambient, gao2020visualechoes} respectively from audio. However, our approach to estimating the geometric cues for audio-visual dereverberation is complementary to these methods. 

\vspace{-0.5mm}
\section{Problem Formulation}

We propose a novel framework that takes reverberant speech $\mathcal{A}_r$ and the corresponding environment panoramic image $\mathcal{V}_r$ as input and outputs estimated clean audio $\mathcal{A}_e$. Both $\mathcal{V}_r$ and $\mathcal{A}_r$ are captured from the listener position focusing on the environment surrounding the speaker (considers far, mid, and near field examples). The reverberation effects can be described using a transfer function known as room impulse response $\mathcal{R}(t)$. $\mathcal{A}_r$ can be obtained by convolving clean speech $\mathcal{A}_s$  with $\mathcal{R}(t)$ (Equation~\ref{rir_eq}) \cite{neely1979invertibility}. Here, $\mathcal{R}$ depends on the listener and speaker positions, room geometry, and acoustic material characteristics. 

\begin{equation}
% $$   
% \belowdisplayskip 0.6\belowdisplayshortskip
\begin{aligned}
\mathcal{A}_r(t) = \mathcal{A}_s(t) * \mathcal{R}(t)
\end{aligned}
\label{rir_eq}
% \vspace{-0.3cm}
% $$
\end{equation}

\section{Our Approach: \modelname}
\label{section: methodology}

\begin{figure*}[h]
    \centering
    \includegraphics[width=0.9\textwidth]{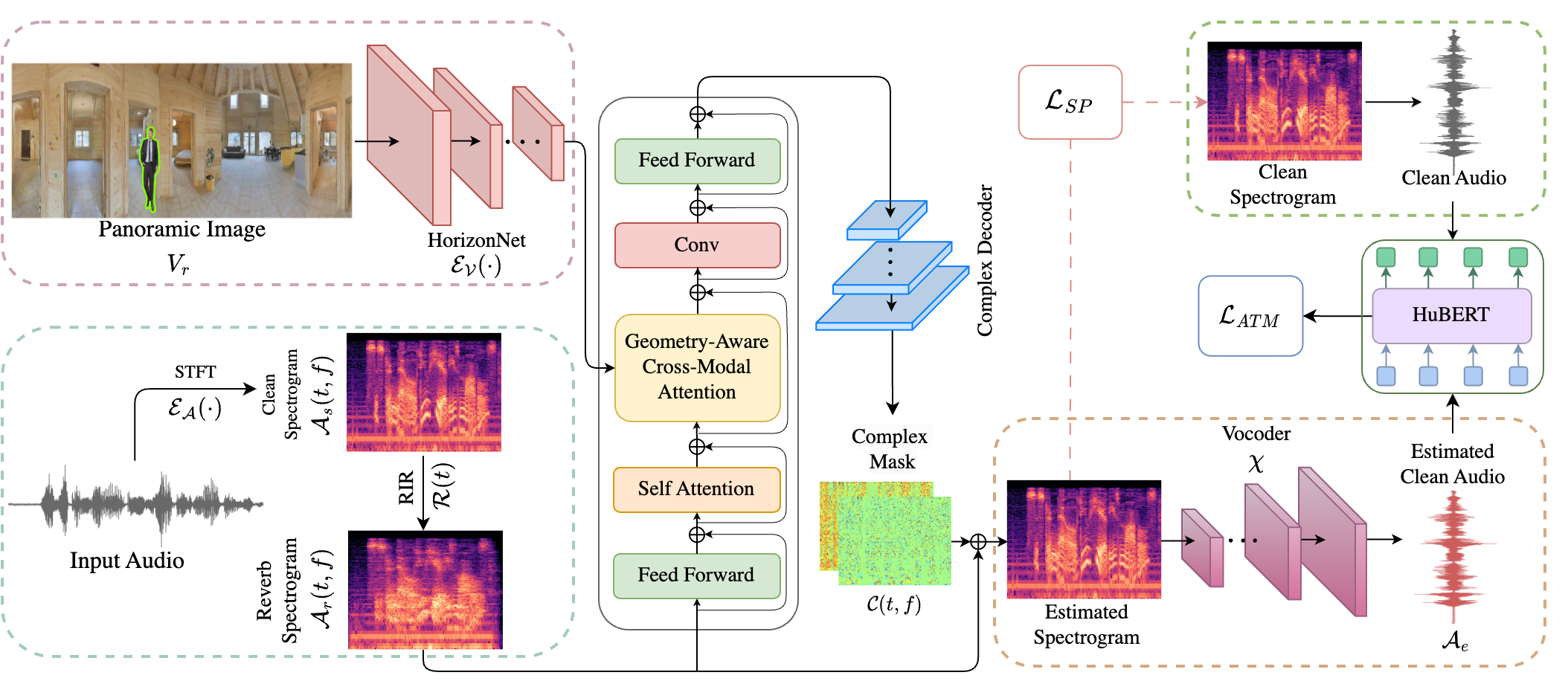}
    \vspace{-2mm}
    \caption{Overview of \textbf{AdVerb}. AdVerb estimates clean source audio from a reverberant speech signal leveraging two primary components: \textcircled{\raisebox{-0.9pt}{1}} The visual stream processing path comprises a HorizonNet-based backbone $\mathcal{E_V(\cdot)}$ to obtain 1D feature sequences, which are subsequently passed to the cross-modal geometry-aware attention subnetwork. \textcircled{\raisebox{-0.9pt}{2}} The audio processing module applies STFT $\mathcal{E_A(\cdot)}$ to get 2D spectrograms which are fed to the cross-modal encoder. The cross-attention subnetwork powered by geometry-aware (Shifted) Window Blocks, Panoptic Blocks, and Relative Position Embedding generates a complex ideal ratio mask.}
    \label{fig:adverb_arch}
    \vspace{-5mm}
\end{figure*}

Fig. \ref{fig:adverb_arch} depicts a pictorial representation of \modelname, our proposed audio-visual dereverberation model. Our primary objective is to learn the inverse function given a reverberant audio signal by exploiting the audio and visual cues. Elaborations on the individual components are as follows:

% To achieve this, we process the visual and audio stream of data to generate a complex ideal ratio mask. We subsequently apply this mask to the source reverberant speech to predict the reverberation-free audio. We optimize our model with two objective functions. 

\subsection{Feature Encoder}
\label{subsec:feature_encoder}

{\noindent \textbf{Vision Encoder:}} To encode geometric layout-specific visual features $\mathcal{E_V(\cdot)}$, we use HorizonNet \cite{sun2019horizonnet}, which is based on ResNet-50 \cite{he2016deep} backbone. HorizonNet takes a panoramic image of the surroundings as input $\mathcal{V}$, with dimensions 512 $\times$ 1024 $\times$ 3. The output is a 2D feature map of 4 different scales. For each feature map, the height is down-sampled, and the width $\mathcal{N}$ is up-sampled to obtain 1D spatial property-infused feature sequences with dimension $\mathbb{R}^{\mathcal{D}/4}$ and connect all the feature maps to obtain $\mathbb{R}^{\mathcal{D}}$, where $\mathcal{D}$ is 1024 in our case. 
\vspace{1mm}

{\noindent \textbf{Audio Encoder:}} For audio features, we employ Short-Time Fourier Transform (STFT) $\mathcal{E_A(\cdot)}$ on the reverberant 1D audio $\mathcal{A}$ to obtain a 2D spectrogram $\mathcal{A}(t,f)$, where $t$ and $f$ index time and frequency, respectively. In contrast to prior work, which learns a convolution network for this transformation \cite{chen2022visual}, we employ STFT with the motivation of using complex masks for learning dereverberation. We calculate STFT with a window of size
400 samples or 25 milliseconds, a hop length of 160 samples or 10 milliseconds, and a 512-point FFT. All our audios are sampled at 16kHz.

\subsection{Complex Ideal Ratio Masks}
\label{subsec:complex}

%mask can be learned for anything.. here the audio should not The intuition behind this is firstly, masks are learnable, and they can leverage the strong association with visual cues. Secondly, 

{\noindent \textbf{Intuition Behind Masks:}} We hypothesize that learning to generate clean anechoic speech in an end-to-end fashion might not be effective owing to the nature of the task. Traditionally, the input audio learns to align to the visual cues, which proves to be effective for the visual acoustic matching~\cite{chen2022visual} task. Similarly, synthesizing speech directly has also seen huge success in audio-visual speech enhancement, and separation \cite{guo2021ad,yang2022audio}, where the visual cues have a high correlation with the contents of the speech, e.g., lip movements. However, the dereverberation algorithm tries to learn an inverse function making the task intrinsically challenging.
% As discussed earlier, audio-visual dereverberation involves considerable intricacies, and the available visual cues have only a modest correlation with the contents of the spoken utterance and only contribute to the nature of corruption that the speech signal has gone through. 
From Fig.~\ref{fig:my_label} it is evident that the same speech content incurs heavy reverberation artifacts when the speaker is far away in a reverberant environment (\textcircled{\raisebox{-0.9pt}{B}}) while the corruption of the speech signal is not significant when the speaker is closer in a relatively less reverberant environment (\textcircled{\raisebox{-0.9pt}{A}}). %\textcolor{red}{ a.k.a (write something about different properties of reverberation)}. 
Thus, we hypothesize such visual cues can be instead used to learn a mask that, when applied to the reverberated speech, suppresses reverberation effects. STFT mask prediction has seen success in the past in a variety of tasks, including source separation \cite{zhao2018sound, gao2019co}, speech enhancement \cite{williamson2016complex}, etc.

\vspace{1mm}
{\noindent \textbf{Complex Ideal Ratio Mask Construction:}} A complex ideal ratio mask (cIRM) \cite{williamson2015complex} is an extension of the conventional ideal ratio mask to process the real and imaginary components of an audio signal separately. The product of cIRM and reverberant speech results in estimated clean speech. It is calculated in the time-frequency (T-F) domain, and thus learning to generate cIRM enhances both the magnitude and phase of reverberant speech, improving overall perceptual speech quality. Given the STFT of reverberant speech, $\mathcal{A}_r(t, f)$, and the cIRM, $\mathcal{C}(t, f)$, clean speech, $\mathcal{A}_s(t, f)$, is computed as follows:

\vspace{-2mm}
\begin{equation}
\mathcal{A}_s(t, f) = \mathcal{C}(t, f) * \mathcal{A}_r(t, f)
\end{equation}

where $t$ and $f$ are index time and frequency respectively. Since the STFT is complex, $*$ indicates complex multiplication. $\mathcal{C}(t, f)$ is computed by dividing the STFT of direct speech, by the STFT of reverberant speech:

% \begin{equation}
% \begin{aligned}
% M(t, f) & =\frac{D(t, f)}{Y(t, f)} \\
% & =\frac{D_r(t, f)+j D_i(t, f)}{Y_r(t, f)+j Y_i(t, f)} \\
% & =\frac{Y_r(t, f) D_r(t, f)+Y_i(t, f) D_i(t, f)}{Y_r^2(t, f)+Y_i^2(t, f)} \\
% & +j \frac{Y_r(t, f) D_i(t, f)-Y_i(t, f) D_r(t, f)}{Y_r^2(t, f)+Y_i^2(t, f)} \\
% \end{aligned}
% \end{equation}

\vspace{-2mm}

\begin{equation}
    \mathcal{A}_s^r(t, f)+j \mathcal{A}_s^i(t, f) = \mathcal{C}(t, f) * \mathcal{A}_r^r(t, f)+j \mathcal{A}_r^i(t, f) \\ 
\end{equation}

\vspace{-2mm}

\begin{equation}
\begin{aligned}
\mathcal{C}(t, f) & = \frac{\mathcal{A}_s^r(t, f)+j \mathcal{A}_s^i(t, f)}{\mathcal{A}_r^r(t, f)+j \mathcal{A}_r^i(t, f)} * \frac{\mathcal{A}_r^r(t, f)-j \mathcal{A}_r^i(t, f)}{\mathcal{A}_r^r(t, f)-j \mathcal{A}_r^i(t, f)} \\
& =\frac{\mathcal{A}_s^r(t, f) \mathcal{A}_r^r(t, f) + \mathcal{A}_s^i(t, f) \mathcal{A}_r^i(t, f) }{\mathcal{A}_r^{r2}(t, f)-\mathcal{A}_r^{i2}(t, f)} \\
& + j\frac{\mathcal{A}_s^i(t, f)\mathcal{A}_r^r(t, f)  - \mathcal{A}_s^r(t, f)\mathcal{A}_r^i(t, f) }{\mathcal{A}_r^{r2}(t, f)-\mathcal{A}_r^{i2}(t, f)} \\
\end{aligned}
\end{equation}

\subsection{Cross-Modal Geometry-Aware Conformer}
\label{subsec:conformer}

{\noindent \textbf{Overview:}} In this module, we aim to learn cross-modal attention between audio and visual features, which enables incorporating fine-grained interactions between them in a geometry-aware fashion. The visual and audio feature maps obtained from corresponding encoders are used as inputs here. The sequence of features from each time step represents a part of the input stream. These sequences are then passed to the conformer-based \cite{gulati2020conformer} cross-modal encoder. For the audio stream, we obtain a complex ideal ratio mask by employing a complex-valued self-attention block. This is then fed into the geometry-aware cross-modal self-attention (GCA) block for audio-visual modeling. We specially design a relative position embedding to encode position-specific information. Finally, the learned representations are passed through a complex-valued decoder to generate the predicted cIRM. We next describe these components in detail.
\vspace{1mm}

{\noindent \textbf{Complex Self-Attention:}} The self-attention mechanism \cite{vaswani2017attention} transforms a sequence into a set of vectors, where each vector is computed as the weighted sum of all other vectors. Here the weights are determined by a learnable function based on the similarities between the input and output. The primary difference between conventional and complex self-attention (CSA) is that the latter operates on complex-valued representations and calculates self-attention separately on the real and imaginary parts. We use CSA instead of vanilla SA layers because of the nature of our input spectrogram. For our implementation, we use Complex-Valued Time-Frequency SA (CTSA) proposed in \cite{kothapally2022complex}, which improves over CSA by accurately modeling inter-dependencies between real and imaginary components of the encoded audio features.
\vspace{1mm}

{\noindent \textbf{Geometry-Aware Cross-Modal Encoder:}} A wealth of studies \cite{chen2022visual, chen2021structure} establish that direct concatenation of cross-modal features \cite{gao20192, owens2018audio} might lead to suboptimal performance. A key observation here is such techniques don't seem suitable in our case, as our application demands more robust reasoning on how different regions of the 3D space contribute to the acoustics differently. For instance, if the sound originates from inside a highly absorptive chamber, less reverberation will be noticeable. In contrast, in the case of a reflective surface, an extended reverberation effect will persist. Hence, it is imperative to attend to image patches to study how they contribute to the overall acoustics.

\begin{figure}
    \centering
    \includegraphics[width=7cm]{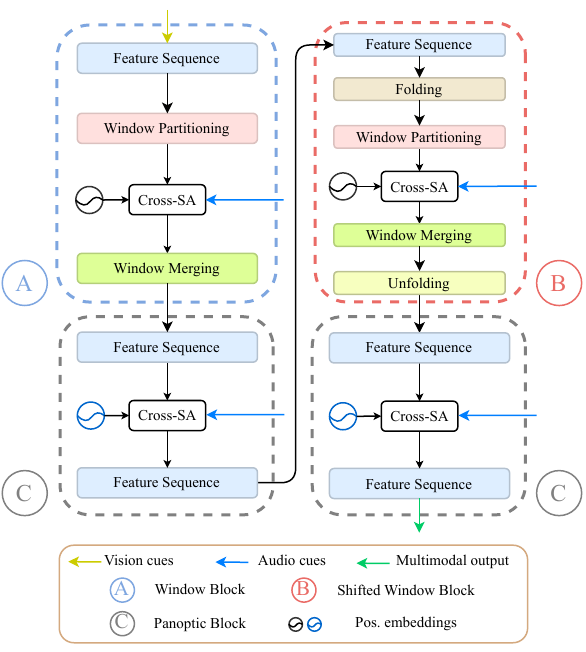}
    \vspace{-2mm}
    \caption{Overview of the Geometry-Aware Cross-Modal Attention block. Window and Panoptic Relative Position Embedding (RPE) are fused into Cross-modal Self-Attention (CSA) blocks. In Window Block \textcircled{\raisebox{-0.9pt}{A}}, partitioning and merging of windows before and after CSA. In \textcircled{\raisebox{-0.9pt}{B}}, Folding and Unfolding of sequence features before and after CSA, respectively. \textcircled{\raisebox{-0.9pt}{C}} integrates another RPE to CSA.}
    \label{fig:cross_modal_attn}
    \vspace{-5mm}
\end{figure}

Inspired by Swin-Transformer \cite{liu2021swin}, our novel GCA module exploits window partitioning for robust spatial modeling ability. However, we observe that using window partition alone limits the conception of the holistic representation of the visual scene. As a result, we equip our Transformer module with (Shifted) Window Blocks and Panoptic Blocks to combine the local and global geometry relations efficiently. Each loop contains four consecutive blocks: Window Block, Panoptic Block, and Shifted Window Block, followed by another Panoptic Block. As shown in Fig.\ref{fig:cross_modal_attn}, the individual blocks follow the Transformer \cite{vaswani2017attention} architecture, with modifications done before and after the multi-head attention layer. Note that the dimension of the sequence and corresponding positions of tokens don't get altered in any block.

In Window Block, we use a patchwise partition on the input feature sequence to obtain $\frac{\mathcal{N}}{\mathcal{N}_w}$ window feature sequences $\mathbb{R}^{\mathcal{N}_w \times \mathcal{D}}$ where $\mathcal{N}_w$ is the window length and is set to 16 in our case. The window partition captures local geometry relations and facilitates the calculation of self-attention by reducing computation while calculating attention. Subsequently, window features are combined after the multi-head attention, as depicted in Fig. \ref{fig:cross_modal_attn} \textcircled{\raisebox{-0.9pt}{A}}.

Inspired by \cite{liu2021swin}, we deploy Shifted Window Block, which connects adjacent windows to facilitate the exchange of information flow between nearby patches. Here a fold and unfold operation is performed by a fraction of $\frac{\mathcal{N}_w}{2}$ to retain the original positions of the feature sequence even after merging: refer to Fig.\ref{fig:cross_modal_attn} \textcircled{\raisebox{-0.9pt}{B}}. Finally, the Panoptic Block follows the native Transformer \cite{vaswani2017attention} encoder to enhance holistic geometry-aware relations of the visual scene (Fig. \ref{fig:cross_modal_attn} \textcircled{\raisebox{-0.9pt}{C}}). 

To model the natural association between visual and audio streams by ensuring cross-modal information flow, we employ the conformer variant \cite{gulati2020conformer} of encoder blocks, which adjoins a convolution layer inside the block for modeling local interactions of audio features. Building on this, we insert one cross-modal attention layer $\xi_{cm}$ after the first feed-forward layer, described as follows:
% $$
% \mathcal{A}_{c m}\left(A_{i}, V_i\right)=\operatorname{softmax}\left(\frac{V_i^{Q} A_i^{K^T{\sqrt{S}}}}\right) V_i^{V}
% $$

\begin{equation}
    \xi_{c m}\left(\mathcal{A}_i, \mathcal{V}_i\right)=\operatorname{softmax}\left(\frac{\mathcal{A}_i^{Q} \mathcal{V}_{i}^{K^T}}{\sqrt{\mathcal{S}}}\right) \mathcal{V}_i^{V}.
\end{equation}

where superscripts $K$, $Q$, and $V$ indicate Key, Query, and Value, respectively. Here, we compute the attention scores between the visual ($\mathcal{V}_i$) and the audio ($\mathcal{A}_i$) sequences by dot-product. This is followed by softmax normalization and scaling by $\frac{1}{\sqrt{\mathcal{S}}}$, which is then used to factor $\mathcal{V}_i$. The key observation here is that cross-modal attention thus designed enables the model to attend to spatial regions in the visual stream and comprehend its acoustic nature.
\vspace{1mm}

{\noindent \textbf{Position Embedding:}}
The conventional attention module is found to be insensitive to the positions of the tokens producing suboptimal results. To this end, we introduce specially designed relative position embedding (RPE) \cite{raffel2020exploring} to strengthen its spatial identification ability. We denote the input sequence of multi-head cross-modal self-attention as $\mathcal{X}=\left\{x_i\right\}_{i=1}^\mathcal{Z}$, where $\mathcal{Z}$ is the sequence length and $x_i \in \mathbb{R}^\mathcal{D}$. A bias matrix $\mathcal{B} \in \mathbb{R}^{\mathcal{Z} \times \mathcal{Z}}$ is added to Scaled Query-Key product \cite{vaswani2017attention}:

\begin{equation}
% $$    
\begin{aligned}
& \alpha_{i j}=\frac{1}{\sqrt{\mathcal{D}}}\left(x_i \mathcal{W}^Q\right)\left(x_j \mathcal{W}^K\right)^T+\mathcal{B}_{i j}, \\
& \text { Attention }(\mathcal{X})=\operatorname{Softmax}(\alpha)\left(\mathcal{X} \mathcal{W}^V\right),
\end{aligned}
% $$
\end{equation}

where $\mathcal{W}^Q, \mathcal{W}^K, \mathcal{W}^V \in \mathbb{R}^{\mathcal{D} \times \mathcal{D}}$ are learnable project matrices and each bias $\mathcal{B}_{i j}$ comes from a learnable scalar table.
In (Shifted) Window Block, $\mathcal{Z}=\mathcal{N}_w$. We denote the learnable scalar table as $\left\{b_k\right\}_{k=-\mathcal{\mathcal{N}}_w+1}^{\mathcal{\mathcal{N}}_w-1}$, and $\mathcal{\mathcal{B}}_{i j}$ corresponds to $b_{j-i}$. This Patch RPE is fed into multi-head attention. 

For Panoptic Block, we consider $\mathcal{Z}=\mathcal{N}$. Here we propose a symmetric representation of only distance and denote the learnable scalar table as $\left\{b_k\right\}_{k=0}^n$, where $n=\frac{\mathcal{N}}{2}$. When $|j-i| \leq \frac{\mathcal{N}}{2}, B_{i j}$ corresponds to $b_{|j-i|}$, otherwise $\mathcal{B}_{i j}$ corresponds to $b_{\mathcal{N}-|j-i|}$.

\subsection{Complex Mask Decoder} 

The complex mask decoder takes input from the conformer and generates a complex ideal ratio mask $\mathcal{C}$. The decoder comprises a complex-valued ReLU activation function followed by a complex-valued convolutional layer, a self-attention module, a dense block, and finally a normalization layer.

\subsection{Vocoder}
After generating the complex ideal ratio mask $\mathcal{C}$, we decode the output spectrogram $\mathcal{G}$ by performing the complex multiplication operation between $\mathcal{C}$ and $\mathcal{G}$. Next, we use a pre-trained vocoder $\chi$ \cite{su2020hifi} to convert the spectral representation of the audio signal to the waveform. We perform this step specifically to calculate the SSL-based HuBERT Loss, which we describe later.

% {\noindent \textbf{Audio Re-synthesis via Vocoder $\mathcal{V}$.}}

\subsection{Model Optimization}
\label{subsec:model_optimization}

% {\noindent \textbf{\novellossone}} To optimize our model, we use the Mean Squared Error(MSE) loss between the predicted spectraogram $\mathcal{S}_g$ and ground-truth spectrogram $\mathcal{S}_p$ as our primary loss function. We define this as:

% \begin{equation}
%     L_{mse}=\left\|\mathcal{S}_p^i-\mathcal{S}_g^i\right\|_2
% \end{equation}

{\noindent \textbf{\novellossone}:} The first of the two objective functions we use for model optimization is the \novellossone ~(SP). Learning to reconstruct the clean spectrogram is a common optimization methodology used in speech enhancement and dereverberation \cite{chen2020learning,kothapally2022skipconvgan}. It computes the $L_{2}$ norm between the spectrogram predicted by our network $\Theta$ and the ground truth clean spectrogram $\mathcal{A}_{s}$. It is defined as: 
\begin{equation}
    \mathcal{L}_{SP}= \mathbb{E}_{(\mathcal{A}_{r}, \mathcal{V}) \sim \mathcal{U}} \left\| \phi(\Theta(\mathcal{A}_{r},\mathcal{V}))-\phi(\mathcal{A}_{s}))\right\|_2,
\end{equation}

where $\mathcal{A}_{r}$ is the reverberant audio and $\mathcal{V}$ is the corresponding panoramic image in some distribution $\mathcal{U}$. $\phi$ is the function that transforms the speech waveform to the corresponding spectrogram representation.
\vspace{1mm}

{\noindent \textbf{\novellosstwo}:} Inspired by the recent success of self-supervised speech representation learning \cite{mohamed2022self}, we introduce \novellosstwo~(ATM). The traditional MSE loss ignores the inherent speech characteristics, like phonetic and prosodic properties, that are essential for learning and reconstructing speech information \cite{hsieh2020improving}. Speech representations learned with SSL effectively encode such characteristics in their latent representations \cite{pasad2021layer}. Thus, we propose a simple yet effective method to enforce the output speech from \modelname~ to encode such information by solving the \novellosstwo~. To calculate ATM loss, we first generate latent representations $\tilde{\mathcal{H}} \in \mathbb{R}^{\mathcal{J}\times d}$ from the clean waveform $\mathcal{A}_s$ with a pre-trained HuBERT \cite{hsu2021hubert} model $\operatorname{e}(\cdot)$, where $d$ is the HuBERT embedding dimension and $\mathcal{J}$ is the sequence length. Next, we cluster these latent representations using \textit{K-means} to generate a sequence of pseudo-labels $\mathcal{P}=\left\{p_t\right\}_{t=1}^\mathcal{J}$. These pseudo-labels are representative of the latent space in our speech input. Finally, we predict these pseudo-labels from latent representations of $\mathcal{A}_e$ (estimated output audio from AdVerb) obtained after passing it through HuBERT. The ATM Loss function can be expressed as follows:

\begin{equation}
    \mathcal{L}_{ATM}(e ; \mathcal{A}_s, \mathcal{A}_e)=\sum_{t \in \mathcal{J}} \log p_f\left(p_t \mid \tilde{\mathcal{H}}, t\right)
\end{equation}

\begin{table*}[]
\centering
\scalebox{0.8}{%
\begin{tabular}{@{}ll|ccccc|c@{}}
% \toprule
\toprule
\multicolumn{2}{l|}{}                                  & \textbf{Speech Enhancement (SE)$^{\dagger}$}              & \multicolumn{2}{c}{\textbf{Speech Recognition (SR)$^{\dagger}$}}                                          & \multicolumn{2}{c|}{\textbf{Speaker Verification (SV)$^{\dagger}$}}                                       & \textbf{RTE* ↓}                          \\
\multicolumn{2}{l|}{\multirow{-2}{*}{\textbf{Method}}} & \textbf{PESQ ↑}                                 & \textbf{WER (\%) ↓}                           & \textbf{WER-FT (\%) ↓}                        & \textbf{EER (\%) ↓}                             & \textbf{EER-FT (\%) ↓}                          & \textbf{(in sec)}                      \\ \midrule
\multicolumn{2}{l|}{Anechoic (Upper bound)}            & 4.72                                          & 2.89                                          & 2.33                                          & 1.53                                          & 1.57                                          & -                                      \\ \midrule
\multicolumn{2}{l|}{Reverberant}                       & 1.49                                          & 8.20                                          & 4.44                                          & 4.51                                          & 4.88                                          & 0.382                                  \\
\multicolumn{2}{l|}{MetricGAN+ \cite{fu2021metricgan+}$^\ddagger$}               & 2.45 (+64\%)                                  & 7.48 (+9\%)                                   & 4.86 (-9\%)                                   & 4.67 (-4\%)                                   & 2.85 (+42\%)                                  & 0.187                                  \\
\multicolumn{2}{l|}{HiFi-GAN \cite{kong2020hifi}$^\ddagger$}                 & 1.83 (+23\%)                                  & 9.31 (-14\%)                                  & 5.59 (-26\%)                                  & 4.32 (+4\%)                                   & 2.49 (+49\%)                                  & 0.196                                  \\
\multicolumn{2}{l|}{WPE \cite{nakatani2010speech}$^\ddagger$}                      & 1.63 (+9\%)                                   & 8.43 (-3\%)                                   & 4.30 (+3\%)                                   & 5.90 (-31\%)                                  & 4.11 (+16\%)                                  & 0.173                                  \\
\multicolumn{2}{l|}{SkipConvGAN \cite{kothapally2022skipconvgan}$^\ddagger$}              & 2.10 (+41\%)                                  & 7.22 (+12\%)                                  & 4.17 (+6\%)                                   & 4.86 (-8\%)                                   & 3.98 (+18\%)                                  & 0.119                                  \\
\multicolumn{2}{l|}{VIDA \cite{chen2021learning}}                     & 2.37 (+59\%)                                  & 4.44 (+46\%)                                  & 3.66 (+18\%)                                  & 3.97 (+12\%)                                  & 2.40 (+51\%)                                  & 0.155                                  \\ \midrule
                     & AdVerb w/o Image                & 2.31 (+55\%)                                  & 3.92 (+52\%)                                  & 3.41 (+23\%)                                  & 3.67 (+19\%)                                  & 2.19 (+55\%)                                  & 0.119                                  \\
                     & AdVerb w/ Random Image          & 2.54 (+70\%)                                  & 4.12 (+50\%)                                  & 3.62 (+18\%)                                  & 3.76 (+17\%)                                  & 2.26 (+54\%)                                  & 0.110                                  \\
                     & AdVerb w/o ATM Loss             & 2.89 (+94\%)                                  & 4.67 (+43\%)                                  & 3.66 (+18\%)                                  & 3.17 (+30\%)                                  & 2.07 (+58\%)                                  & 0.117                                  \\
                     & AdVerb w/o Complex SA           & 2.91 (+95\%)                                  & 3.63 (+56\%)                                  & 2.98 (+33\%)                                  & 3.21 (+29\%)                                  & 2.10 (+57\%)                                  & 0.117                                  \\
                     & AdVerb w/o Geometry Aware Block & 2.30 (+54\%)                                  & 4.01 (+51\%)                                  & 3.12 (+30\%)                                  & 3.68 (+18\%)                                  & 2.12 (+57\%)                                  & 0.113                                  \\
                     & AdVerb w/o RPE                  & 2.79 (+87\%)                                  & 3.54 (+57\%)                                  & 3.01 (+32\%)                                  & 3.17 (+30\%)                                  & 2.11 (+57\%)                                  & 0.107                                  \\
                     & AdVerb w/o Window Block         & 2.81 (+89\%)                                  & 3.61 (+56\%)                                  & 2.99 (+33\%)                                  & 3.14 (+30\%)                                  & 2.12 (+57\%)                                  & 0.108                                  \\
\multirow{-8}{*}{\begin{turn}{90}\textbf{ABLATION}\end{turn} }  & AdVerb w/o Panoptic Block       & 2.92 (+96\%)                                  & 3.59 (+56\%)                                  & 2.92 (+34\%)                                  & 3.29 (+27\%)                                  & 2.01 (+59\%)                                  & 0.107                                  \\ \midrule
\multicolumn{2}{l|}{\textbf{AdVerb (ours)}}            & \cellcolor[HTML]{E2FEFF}\textbf{2.96 (+98\%)} & \cellcolor[HTML]{E2FEFF}\textbf{3.54 (+57\%)} & \cellcolor[HTML]{E2FEFF}\textbf{2.91 (+34\%)} & \cellcolor[HTML]{E2FEFF}\textbf{3.11 (+31\%)} & \cellcolor[HTML]{E2FEFF}\textbf{1.98 (+59\%)} & \cellcolor[HTML]{E2FEFF}\textbf{0.101} \\ \bottomrule

% \multicolumn{2}{l|}{\textbf{AdVerb (ours)}}        & \cellcolor[HTML]{E2FEFF}{{\color[HTML]{333333} \textbf{2.96 (+98\%)}}}                           & \cellcolor[HTML]{E2FEFF}{{\color[HTML]{333333} \textbf{3.54 (+57\%)}}}  & \cellcolor[HTML]{E2FEFF}{\textbf{2.91 (+34\%)}}  & {\color[HTML]{333333} \cellcolor[HTML]{E2FEFF}{\textbf{ 3.11 (+31\%)}}}   & \cellcolor[HTML]{E2FEFF}{\textbf{1.98 (+59\%)}} & \cellcolor[HTML]{E2FEFF}\textbf{0.101}
\end{tabular}%
}
\vspace{-2mm}
\caption{Comparison of AdVerb with various baselines on multiple spoken language processing tasks based on the LibriSpeech test-clean set (marked with $\dagger$) and on sim-to-real transfer based on the AVSpeech dataset (marked with *). “Anechoic (Upper bound)" refers to clean speech, while ``Reverberant'' refers to clean speech convolved with RIR. WER-FT and EER-FT denote evaluations when the SR and SV models are finetuned with the audio-enhanced data. Numbers in parentheses denote the relative improvement compared to Reverberant. Methods marked with $\ddagger$ are audio-only.}
\label{tab:table1_iclr21}
\vspace{-3mm}
\end{table*}
\vspace{4mm}

% {\noindent \textbf{Total loss}}

\noindent where $p_f$ is the distribution over the target indices at each timestep $t$. Finally, we optimize our model with a total loss $\mathcal{L}$ as follows:
\begin{equation}
\mathcal{L}=\lambda \mathcal{L}_{SP}+\mu \mathcal{L}_{ATM}
\end{equation}
where $\lambda, \mu \in  \mathbb{R}$ are hyper-parameters to balance the contribution of each loss component.

% Image view rotation was used as a part of data augmentation. This is possible because our audio recording is omnidirectional and is independent of the camera pose. This data augmentation strategy prevents the model from overfitting; without it our model fails to converge. This strategy creates a one-to-many mapping between reverb and views, forcing the model to learn a viewpoint-invariant representation of the room acoustics

\section{Experiments and Results}
For a fair assessment of our model, we evaluate our model through speech dereverberation on three downstream tasks: speech enhancement (SE), automatic speech recognition (ASR), and speaker verification (SV), respectively. The environments are taken from Matterport3D \cite{chang2017matterport3d}, with speech samples from the LibriSpeech dataset \cite{panayotov2015librispeech}.

\subsection{Dataset}
\noindent \textbf{SoundSpaces-Speech Dataset:} We use the SoundSpaces-Speech dataset proposed in \cite{chen2021learning} for our experiments. It comes with paired anechoic and reverberant audio with camera views from 82 Matterport3D \cite{chang2017matterport3d} environment convolved with speech clips from LibriSpeech \cite{panayotov2015librispeech} samples. SoundSpaces \cite{chen2020soundspaces} provide precomputed RIRs $\mathcal{R}(t)$, which are convolved with speech waveforms to obtain reverberant signal $\mathcal{A}_r(t)$ for a total of 49,430/2,700/2,600 train/validation/test samples, respectively.  

\vspace{1mm}
\noindent \textbf{Acoustic AVSpeech Web Videos:} Web videos offer natural supervision between visuals and acoustics in abundance. To be consistent with prior work, we use the collection from \cite{chen2022visual}, which is a subset of the AVSpeech\cite{ephrat2018looking} dataset. The clip durations range between 3-10 seconds with a visible human subject in each video frame. To evaluate our model on real-world data in addition to synthetic data, we use these 3K samples only for testing purposes.

\vspace{1mm}
\noindent \textbf{Evaluation Tasks And Metrics:} We follow the standard practice of reporting Perceptual Evaluation of Speech Quality (PESQ)~\cite{rix2001perceptual}, Word Error Rate (WER), and Equal Error Rate (EER) to compare our method with the baselines for the three tasks. Following \cite{chen2021learning}, we employ the pre-trained models from the SpeechBrain \cite{ravanelli2021speechbrain} for ASR and SV tasks. These models were evaluated on the LibriSpeech test-clean set. SV evaluation was done on a set of 80K randomly sampled utterance pairs from the test-clean set.

\subsection{Baselines}
\label{subsec:baselines}
{\noindent \textbf{WPE}~\cite{nakatani2010speech}:} A statistical method that estimates an inverse system for late reverberation. It deploys variance normalization to improve dereverberation results with relatively short observations.  

\vspace{1mm}
{\noindent \textbf{MetricGan+}~\cite{fu2021metricgan+}}: We use the implementation by \cite{ravanelli2021speechbrain} for benchmarking. As presented by the authors, it can be used to optimize different metrics. We optimize PESQ to report values from the best model for individual downstream tasks. 

\vspace{1mm}
{\noindent \textbf{SkipConvGAN}~\cite{kothapally2022skipconvgan}:}  A recent model where the generator network estimates a complex time-frequency mask and the discriminator aids in driving the generator to restore the lost formant structure. The model achieves SOTA results on the dereverberation task.

\vspace{1mm}
{\noindent \textbf{HiFi-GAN}~\cite{kong2020hifi}:} A GAN-based high-fidelity speech synthesis system that shows satisfactory results on speech dereverberation. It models periodic patterns of audio to enhance sample quality.

\vspace{1mm}
{\noindent \textbf{VIDA}~\cite{chen2021learning}:} An end-to-end vision backed speech dereverberation framework. It combines RGB-D image information to estimate clean speech.  
\vspace{1mm}

\subsection{Results}
\label{subsec:results}
{\noindent \textbf{Evaluation Setup on LibriSpeech:}} We compare model performance on three speech tasks: Speech Enhancement (SE), Automatic Speech Recognition (ASR), and Speaker Verification (SV). 
% We report the Perceptual Evaluation of Speech Quality (PESQ) \cite{rix2001perceptual}, Word Error Rate (WER), and Equal Error Rate (EER) for the three tasks, respectively. 
To evaluate our trained models, we use the dereverbed version of the test-clean set split of the LibriSpeech dataset. Similar to \cite{chen2021learning}, for SR and SV, we either use pre-trained models from SpeechBrain \cite{ravanelli2021speechbrain} or fine-tune a model from scratch using dereverbed LibriSpeech train-clean-360 split.
\vspace{2mm}

{\noindent \textbf{Quantitative Analysis on LibriSpeech:}} Table \ref{tab:table1_iclr21} compares the performance of AdVerb with the baselines. Experimental results show AdVerb outperforms all audio-only baselines by a significant margin on all three tasks. We achieved relative improvements of 41\%, 51\%, and 36\% over the best audio-only baseline, SkipConvGAN, on SE, SR, and SV, respectively, in terms of relative gain from reverberant speech. AdVerb also outperforms VIDA by 25\%, 20\%, and 22\% on SE, SR, and SV, respectively, which shows the superiority of AdVerb in audio-visual dereverberation tasks.
\vspace{1.5mm}

{\noindent \textbf{Quantitative Analysis on AVSpeech:}} To examine the robustness of our proposed approach in real-world settings, we evaluate our model on an in-the-wild AVSpeech audio-visual dataset collected from YouTube~\cite{ephrat2018looking}. The AVSpeech dataset has non-panoramic images; therefore the field-of-view is limited in the test dataset, and the performance of our network trained on panoramic images is not optimal. In the absence of the ground truth clean speech, we use the average reverberation time (RT) of the dereverberated speech signal for evaluation. RT is the time taken to decay the sound pressure in RIR by 60 decibels. We can estimate RT from the reverberant speech signal~\cite{chen2022visual}. According to Equation~\ref{rir_eq}, in clean speech, RIR will be an impulse response ($\delta(t)$) and $\approx$ $0$. The dereverberated speech with the least amount of reverberation will have the lowest RT. Therefore, reverberation time error (RTE) is the average RT of the dereverberated test speech samples. From Table~\ref{tab:table1_iclr21}, we can see that AdVerb reports the lowest RTE.
\vspace{1.5mm}

{\noindent \textbf{Ablation Study:}} To show the importance of the individual components in AdVerb, we perform an extensive ablation study shown in Table \ref{tab:table1_iclr21}. Note that AdVerb sees the steepest fall in performance across tasks when trained and evaluated w/o images, i.e., in an audio-only setup. In this setup, our GCA block is replaced with a simple uni-modal self-attention block. There is also a considerable drop in performance across all tasks w/o the geometry-aware module, thus underlining the importance of this block. In this setup, our GCA block is replaced with a simple cross-modal self-attention block with queries as audio cues and keys and values as visual cues. We carry out further ablations to study the contributions of the individual components of the cross-modal geometry-aware attention block. Interestingly, the drop in performance when removing the individual elements is much less than the entire GCA block. This underlines that these components combine to have a telling impact on the overall setup. Finally, we show that ATM loss improves AdVerb's SR performance by a significant margin. Refer to the supplementary for further ablations. 
\vspace{1.5mm}

{\noindent \textbf{More Comparison Against Audio-only Methods:}} Table \ref{tab:comp_oth} demonstrates the performance of our model against SOTA audio-only methods. \modelname~ outperforms existing audio-only methods and sets new benchmarks.

\begin{table}[h]
\vspace{-2mm}
 \setlength{\tabcolsep}{1.4pt}
\centering
\scalebox{0.6}{
\begin{tabular}{@{}l|c|cc|cc|c@{}}
% \toprule
\toprule
{\color[HTML]{333333} \textbf{}} & \multicolumn{1}{c|}{{\color[HTML]{333333} \textbf{SE}}} & \multicolumn{2}{c|}{{\color[HTML]{333333} \textbf{SR}}} & \multicolumn{2}{c|}{{\color[HTML]{333333} \textbf{SV}}} & \multicolumn{1}{c}{{\color[HTML]{333333} \textbf{\  RTE ↓}}} \\
{\color[HTML]{333333} \textbf{Method}}        & {\color[HTML]{333333} \textbf{PESQ ↑}}                                 & {\color[HTML]{333333} \textbf{WER (\%) ↓}}    & \textbf{WER-FT (\%) ↓} & {\color[HTML]{333333} \textbf{EER (\%) ↓}}     & \textbf{EER-FT (\%) ↓}  & \textbf{(in sec)} \\ \midrule 

{\color[HTML]{333333} Reverberant}            & {\color[HTML]{333333} 1.49}                                            & {\color[HTML]{333333} 8.20}                   & 4.44                   & {\color[HTML]{333333} 4.51}                    & 4.88       & 0.382            \\
{\color[HTML]{333333} DEMUCS \cite{demucs}}   & {\color[HTML]{333333} 2.17 }                                    & {\color[HTML]{333333} 7.97 (+2.8\%) }           & 5.20 (-17\%)         & {\color[HTML]{333333} 3.82 (+15\%) }           & 2.96 (+39\%)       & 0.129      \\ 
{\color[HTML]{333333} VoiceFixer \cite{voicefixer}}   & {\color[HTML]{333333} 2.41}                                    & {\color[HTML]{333333} 5.66 (+31\%)}           & 4.19 (+5\%)            & {\color[HTML]{333333} 3.76 (+16\%) }           & 2.79 (+42\%)      & 0.121       \\
{\color[HTML]{333333} H-GAN \cite{h-gan}}   & {\color[HTML]{333333} 1.94}                                    & {\color[HTML]{333333} 8.14 (+1\%) }           & 5.01 (-12\%)          & {\color[HTML]{333333} 4.22 (+6\%)}           & 3.13 (+35\%)       & 0.196      \\
{\color[HTML]{333333} Kotha. \textit{et al.} \cite{kothapally2022complex}}   & {\color[HTML]{333333} 2.54 }                                    & {\color[HTML]{333333} 5.32 (+35\%) }           & 4.13 (+6\%)            & {\color[HTML]{333333} 3.71 (+17\%)}           & 2.68 (+44\%)      & 0.124       \\
\midrule
{\color[HTML]{333333} \textbf{AdVerb}}   & \cellcolor[HTML]{E2FEFF}{\color[HTML]{333333} \textbf{2.96}}                & \cellcolor[HTML]{E2FEFF}{\color[HTML]{333333} \textbf{3.54 (+57\%)}} & \cellcolor[HTML]{E2FEFF}{\textbf{2.91 (+34\%)}}           & \cellcolor[HTML]{E2FEFF}{\color[HTML]{333333} \textbf{3.11 (+31\%)}}           & \cellcolor[HTML]{E2FEFF}\textbf{1.98 (+59\%)}      & \cellcolor[HTML]{E2FEFF}{\textbf{0.101}}       \\
 \bottomrule
\end{tabular}}
% \vspace{-2mm}
\caption{Comparison of AdVerb with more audio-only approaches. AdVerb results in a relative gain of \textbf{14\%-56\%}. Percentages in bracket represent an improvement on reverberant audio.}
\label{tab:comp_oth}
\vspace{-2mm}
\end{table}
\vspace{1.5mm}

\noindent \textbf{Results on Noisy Dataset}: To evaluate the robustness of the proposed AdVerb model to outdoor unwanted noise, we add ambient sounds from urban environments to the LibriSpeech test-clean dataset using the MUSAN dataset \cite{snyder2015musan}. Following \cite{chen2020learning}, we maintain an SNR of 20 for our mixture. Table \ref{tab:noisy1} compares the performance of AdVerb on three downstream speech-based tasks. All experiments were done for the non-fine-tuned version of our experimental setup, where a pre-trained model was used from SpeechBrain. Though we see a drop in performance compared to the noise-free dataset, AdVerb outperforms all our baselines and maintains similar margins compared to the original noise-free dataset.

\begin{table}[h]
\centering
\scalebox{0.86}{
\begin{tabular}{@{}l|cccc}
% \toprule
\toprule
 & \textbf{SE} & \textbf{SR} & \textbf{SV}     \\
  \textbf{Method} & \textbf{PESQ ↑} & \textbf{WER ↓} & \textbf{EER ↓} \\
\midrule
Anechoic (Upper bound)  & 4.72    & 2.89     & 1.53  \\\hline
Reverberant   & 1.57   & 11.45     & 4.76  \\
MetricGAN+   & 2.29   & 8.92     & 4.89  \\
HiFi-GAN & 1.95   & 10.55    & 4.73  \\
WPE  & 1.88   & 9.10     & 5.11   \\ 
SkipConvGAN  &  2.06  & 7.28    & 4.94  \\ 
VIDA  &  2.14   & 4.97    & 4.01  \\ \bottomrule
\textbf{AdVerb (Ours)}  & \cellcolor[HTML]{E2FEFF}\textbf{2.52}  & \cellcolor[HTML]{E2FEFF}\textbf{4.20}    & \cellcolor[HTML]{E2FEFF}\textbf{3.46}  \\
\bottomrule
\end{tabular}
}
% \vspace{-2mm}
\caption{Result comparison of AdVerb with baseline methods on noise added dataset splits for 3 speech tasks.}
\label{tab:noisy1}
\vspace{-4mm}
\end{table}

\noindent \textbf{Ablation on Noisy Dataset}: Table \ref{tab:noisy2} illustrates the results of the ablation study on the noisy LibriSpeech dataset. The noise addition process is the same as before.

\begin{table}[h]
\vspace{-2mm}
\centering
\scalebox{0.88}{
\begin{tabular}{@{}l|cccc}
% \toprule
\toprule
 & \textbf{SE} & \textbf{SR} & \textbf{SV}     \\
  \textbf{Method} & \textbf{PESQ ↑} & \textbf{WER ↓} & \textbf{EER ↓} \\
\midrule
% Anechoic (Upper bound)  & 4.72    & 2.89     & 1.53  \\\hline
w/o Image   & 2.03  & 4.68     & 3.81  \\
w/o ATML   & 2.28   & 5.10     & 3.87 \\
w/o Geom. aware & 2.29   & 4.99    & 3.64  \\
w/o Window block & 2.34   & 4.43    & 3.43   \\ 
w/o Panoptic block &  2.39  & 4.37    & 3.51  \\  
% \textbf{AdVerb (Ours)}  & \cellcolor[HTML]{E2FEFF}\textbf{2.52}  & \cellcolor[HTML]{E2FEFF}\textbf{4.20}    & \cellcolor[HTML]{E2FEFF}\textbf{3.46}  \\
\bottomrule
\end{tabular}}
% \vspace{1mm}
\caption{Ablation on LibriSpeech noisy data. AdVerb performs considerably well on noisy data with the individual modules contributing to the overall gain.}
\label{tab:noisy2}
\vspace{-5mm}
\end{table}

\begin{figure*}
     \centering
     \begin{subfigure}[b]{0.48\textwidth}
         \centering
         \includegraphics[width=\textwidth]{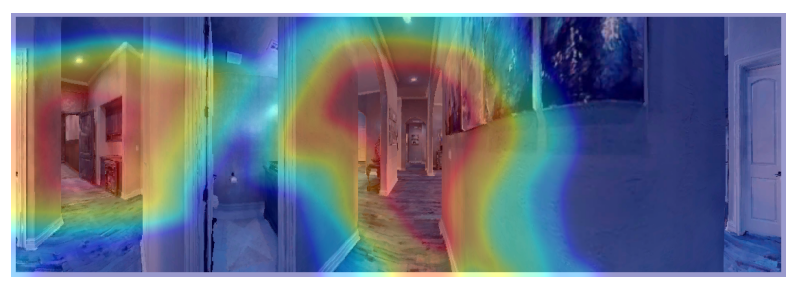}         
         \vspace{-6mm}
         % \caption{Atrium}
         \label{fig:atrium2}
     \end{subfigure}
     \hfill
     \begin{subfigure}[b]{0.48\textwidth}
         \centering
         \includegraphics[width=\textwidth]{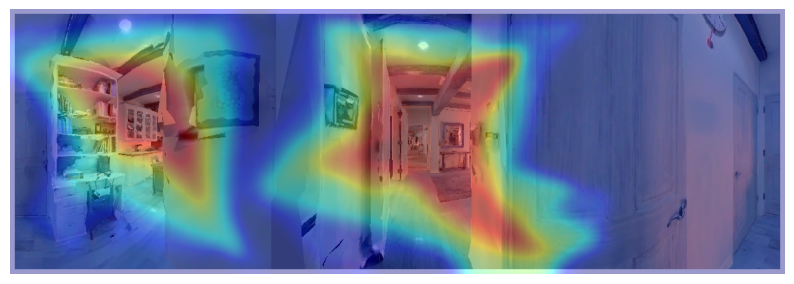}
         \vspace{-6mm}
         % \caption{Long Corridor}
         % \label{fig:corridor}
     \end{subfigure}
     \\
     \begin{subfigure}[b]{0.48\textwidth}
         \centering
         \includegraphics[width=\textwidth]{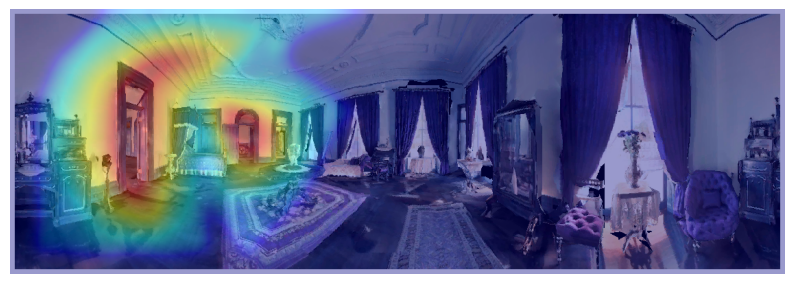}
         \vspace{-6mm}
         % \caption{Successful cases.}
         % \label{fig:library11}
     \end{subfigure}
     \hfill
     \begin{subfigure}[b]{0.48\textwidth}
         \centering
         \includegraphics[width=\textwidth]{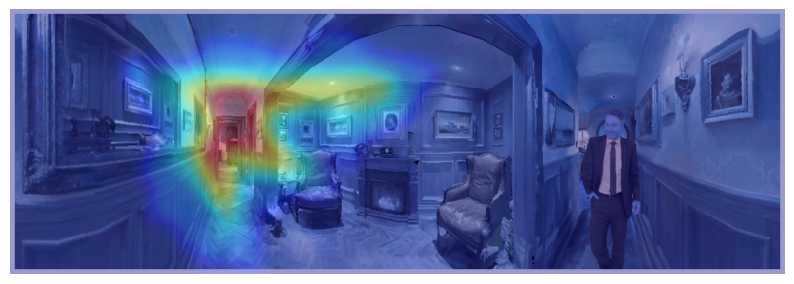}
         \vspace{-6mm}
         % \caption{Successful cases.}
         % \label{fig:library}
     \end{subfigure}
    \caption{Grad-CAM visualization of activated regions. Our model attends to regions that cause heavy reverberation effects.}
        \label{fig:grad_cam}
\end{figure*}

\begin{figure*}
\centering
    \begin{subfigure}[b]{0.48\textwidth}
         \centering
         \includegraphics[width=\textwidth]{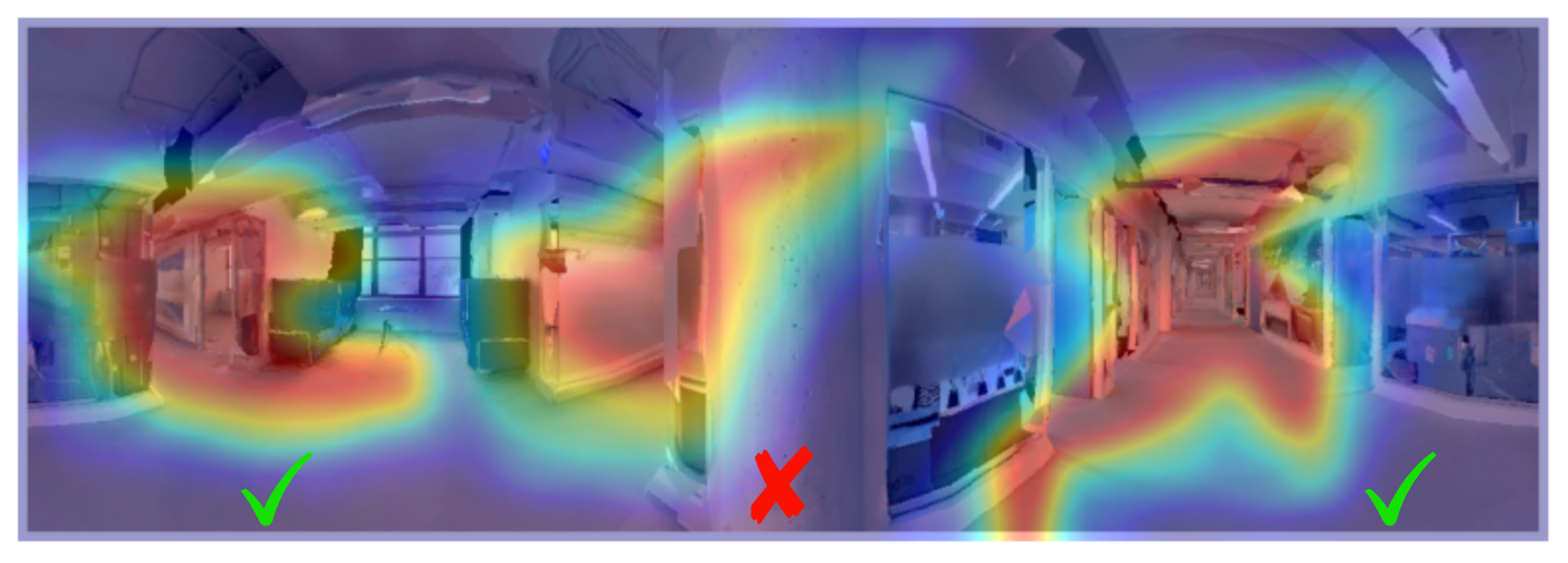}
         \vspace{-6mm}
         % \caption{Meeting room}
         % \label{fig:meeting_room}
     \end{subfigure}
     \hfill
     \begin{subfigure}[b]{0.48\textwidth}
         \centering
         \includegraphics[width=\textwidth]{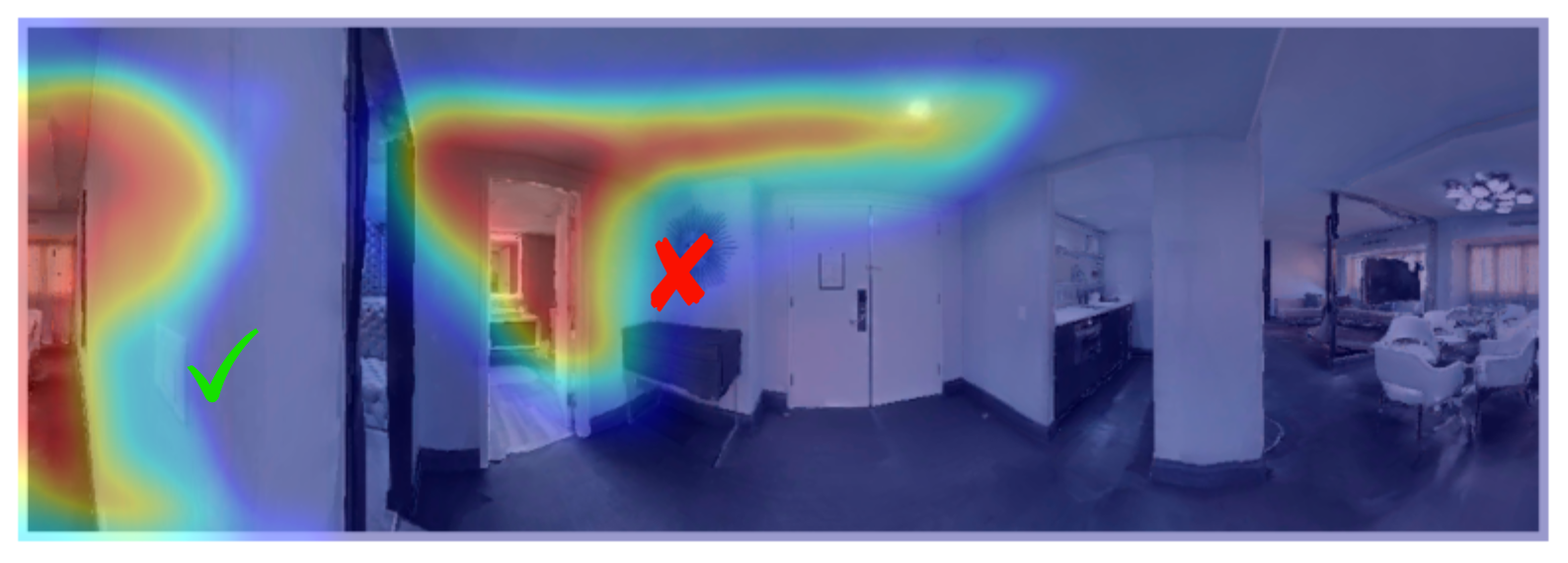}
         \vspace{-6mm}
         % \caption{Meeting room}
         \label{fig:meeting_room}
     \end{subfigure}
        \caption{Some failure cases. The \color{green}{\checkmark} \color{black} denotes the regions with correct activation while \color{red}{\xmark} \color{black} spurious detections.} 
        \label{fig:grad_cam_tick}
\end{figure*}

% \begin{figure}
%     \centering
%     \includegraphics[width=8cm]{figs/attn_map.png}
%     \vspace{-4mm}
%     \caption{Grad-CAM activation of a hallway scene. The corridor regions are more activated as they cause more reverberation.}
%     \label{fig:attn_map}
%     \vspace{-6mm}
% \end{figure}

% \noindent \textbf{Performance under noisy environment:}
\vspace{1.5mm}
\noindent \textbf{Analysis On Visual Features:}
To underline the importance of the visual cues, we show the activations of the network using Grad-CAM\cite{selvaraju2017grad} in Fig. \ref{fig:grad_cam}. Note that the network attends to the sides of the hallway or empty regions with almost or no sound absorbers which lead to longer reverberation effects. Fig. \ref{fig:grad_cam_tick}
demonstrates some cases where our model attends to spurious regions.

% \usepackage{booktabs}
\begin{table}[]
\centering
 \setlength{\tabcolsep}{1.4pt}
\scalebox{0.9}{
\begin{tabular}{@{}l|c|l@{}}
% \toprule
\toprule
\textbf{Baseline}   & \textbf{SoundSpaces(in \%)} & \textbf{AVSpeech(in \%)} \\
\textbf{Method}   & \textbf{(A\% / B\% / C\%)}& \textbf{(A\% / B\% / C\%)}\\ \midrule
Clean Speech     &  \textbf{61.3} / 8.1 / 30.6              &  \multicolumn{1}{c}{-- \hfill/\hfill -- \hfill/\hfill --}            \\
VIDA\cite{chen2021learning}   & 16.5 / 6.5 / \textbf{\underline{77.0}}               & \hfill 13.5 / 0.0 / \textbf{\underline{86.5}}               \\
WPE\cite{nakatani2010speech}          & 8.8 / 3.5 / \textbf{\underline{87.7}}                & \hfill 3.7 / 7.4 / \textbf{\underline{88.9}}              \\
SCGAN\cite{kothapally2022skipconvgan} & 9.2 / 0.0 / \textbf{\underline{90.8}}                & \hfill 0.0 / 8.0 / \textbf{\underline{92.0}}             \\ \bottomrule
\end{tabular}}
\vspace{1mm}
\vspace{-2mm}
\caption{User study results. A\% of participants find the baseline audio samples better, B\% have no preference, and C\% prefer \modelname.}
\label{tab:user_study1}
\vspace{-5mm}
\end{table}

% \vspace{-1mm}
\subsection{User Study For Subjective Evaluation}
In addition to objective metric evaluation, we perform a subjective human listening study on a synthetic (generated using SoundSpaces) and an in-the-wild (AVSpeech) dataset over Amazon MTurk. We believe this can be a good measure to understand how realistic and aesthetically pleasing the output produced by our model is. Moreover, through this, we try to understand other aural artifacts not captured in an objective measure like PESQ. In our study, a total of 89 participants were presented with 8 sets of samples containing the reverberant speech, clean speech (not present for AVSpeech), and estimated dereverberant speech. Table \ref{tab:user_study1} demonstrates that users find samples generated by our method better than the three other baselines VIDA \cite{chen2021learning}, WPE \cite{nakatani2010speech} and SkipConvGAN \cite{kothapally2022skipconvgan}, in both cases.

\vspace{-1mm}

\section{Conclusions and Future Works}
In this paper, we present a novel audio-visual dereverberation framework. To this end, we introduce the GCA module with a specially designed position embedding scheme to capture the local and global spatial relations of the 3D environment. The experimental analysis demonstrates how modeling the visual information efficiently can lead to improved performance of such a system. We believe our work will encourage further research in this space. One limitation of our approach is that the efficacy of the method drops for non-panoramic images. Future work can aim towards finding more sophisticated ways of modeling the acoustic property of the environment and combining cross-modal information. Although our framework achieves highly satisfactory results at all difficulty levels on both simulated and real-world samples, we notice the performance of our model can be improved for situations with extreme reverb effects, and far away subjects. A potential use case of our work can be to leverage the properties of target visual scenes to provide immersive experiences to users in AR/VR applications. This work can also find applications in the audio/speech simulation domain.

% \balance
{\small
\bibliographystyle{ieee_fullname}
\bibliography{final}
}

\pagebreak

\section{Appendix}

% In this supplemental, we provide the following additional
% material to the main submission:

% \begin{enumerate}   %[label=(\Alph*)]
%     \item Estimated Dereverbed Audio Samples
%     \item Additional Experiments And Results
        
%         \begin{enumerate}
%     	  \item Results on Noisy Data
%             \item Performance with 3D Humanoid Removed
%             % \item Environment Type and Speaker Distance
            
%         \end{enumerate}
%     \item Further Ablations
%         \begin{enumerate}
%             \item Role of Complex Ideal Ratio Mask
%         \end{enumerate}
%     \item Implementation Details
%     \item Evaluation Details
%     \item Hyperparameter Tuning Experiments

%     \item User Study 
%     \item Visual Environments and Sample Curation
%     \item Societal Impact
%     \item Limitations
% \end{enumerate}

% \hspace{2 cm}

\subsection{Estimated Dereverbed Audio Samples}

We provide audio samples of the estimated dereverberant audio generated by \modelname ~ in the project webpage (use of headphones recommended). Some of these samples were used during the user study. We observe that the perceptual quality of the audio samples generated by our model is significantly better compared to other prior methods (established by the user study).

\subsection{Additional Experiments And Results}

% by adding ambient sounds from urban environments such as
% coffee shops, restaurants, and bars using the WHAM dataset [64]. We add them to the reverberant
% test waveform with a SNR of 20, following [14, 45]. Table 4 shows the results on three downstream
% tasks. As expected, all models’ performance drop compared to the results in the noise-free test setting
% (Table 1), but our VIDA model still significantly outperforms the baselines on ASR. For speaker
% verification, WPE [45] is reported to be robust to noisy input and thus has lower EER while using
% the pretrained model, but it underperforms VIDA when the SV model is finetuned on the enhanced
% speech. Noise has less impact on the performance on MetricGAN+[16] likely because it directly
% optimizes PESQ.

% \subsubsection{Role of Window Size}

% In order to reduce computation and enhance geometry-aware modeling ability we employ window partition. However, experimental results show window partition alone leads to lower global modeling ability. Thus, we need to equip the Transformer of (Shifted) Window Blocks and Panoptic Blocks to combine the local and global geometry relations. 

% \subsubsection{Environment Type and Speaker Distance}

\subsubsection{Performance with 3D Humanoid Removed}

To inspect the influence of human speaker cues, we carry out an experiment with the humanoid removed keeping everything else the same. We observe that all the metrics become impacted and achieve inferior scores (Table \ref{tab:humanoid}). We conclude the speaker's location-specific information is crucial for better learning ability of the model in order to perform dereverberation. 

\begin{table}[h]
\centering
\begin{tabular}{@{}l|cccc}
% \toprule
\toprule
 & \textbf{SE} & \textbf{SR} & \textbf{SV}     \\
 & \textbf{PESQ ↑} & \textbf{WER ↓} & \textbf{EER ↓} \\
\midrule
AdVerb \textit{w/o human mesh}  & 2.94  & 3.67   & 3.15  \\
AdVerb \textit{w/ human mesh}  & 2.96  & 3.54    & 3.11  \\ \bottomrule
\end{tabular}
\vspace{1mm}
\caption{Result comparison of AdVerb with and without 3D humanoid in the panoramic images.}
\label{tab:humanoid}
% \vspace{-2mm}
\end{table}

\subsection{Further Ablations on LibriSpeech dataset}

\subsubsection{Role of Complex Ideal Ratio Mask}
As described in Section 4.4, the complex Ideal Ratio Mask (cIRM) is an extension of the conventional ideal ratio mask (IRM) to process the real and imaginary components of the STFT separately. Thus, to measure the influence of the complex aspect of cIRM, we replace it with IRM and make the necessary updates in the pipeline. Table \ref{tab:cirm} compares the performance of both models. As can be observed, employing cIRM improves performance over using the conventional mask-based pipeline, which is consistent with prior findings \cite{williamson2016complex}.

\begin{table}[h]
\centering
\begin{tabular}{@{}l|cccc}
% \toprule
\toprule
 & \textbf{SE} & \textbf{SR} & \textbf{SV}     \\
 & \textbf{PESQ ↑} & \textbf{WER ↓} & \textbf{EER ↓} \\
\midrule
AdVerb \textit{w/ IRM}  & 2.52  & 4.12   & 3.85  \\
AdVerb \textit{w/ CIRM}  & 2.96  & 3.54    & 3.11  \\ \bottomrule
\end{tabular}
\vspace{1mm}
\caption{Result comparison of AdVerb with and without the Complex Ideal Ratio Mask.}
\label{tab:cirm}
\vspace{-2mm}
\end{table}

% \subsection{Implementation Details}
% {\noindent \textbf{Audio STFT.}} We calculate STFT with a window of size
% 400 samples or 25 milliseconds, a hop length of 160 samples or 10 milliseconds, and a 512-point FFT. All our audios are sampled at 16kHz.

\begin{figure*}
     \centering
     \begin{subfigure}[b]{0.48\textwidth}
         \centering
         \includegraphics[width=\textwidth]{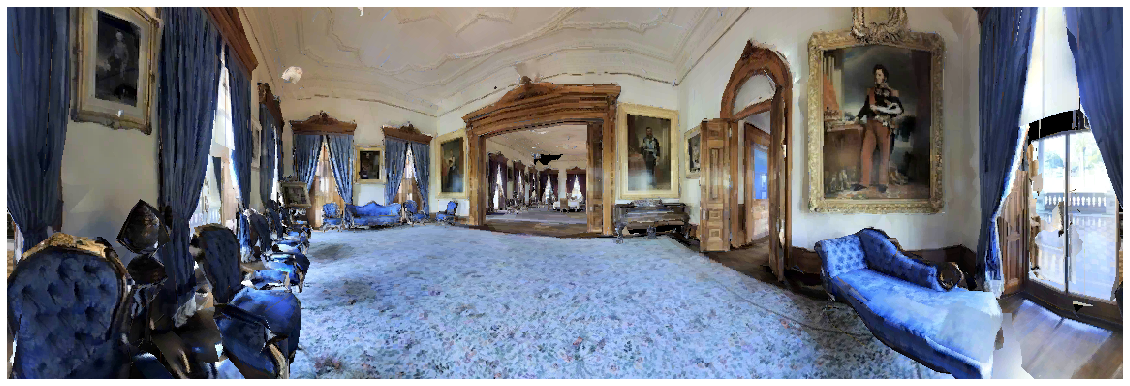}         
         \vspace{-6mm}
         \caption{Atrium}
         \label{fig:atrium1}
     \end{subfigure}
     \hfill
     \begin{subfigure}[b]{0.48\textwidth}
         \centering
         \includegraphics[width=\textwidth]{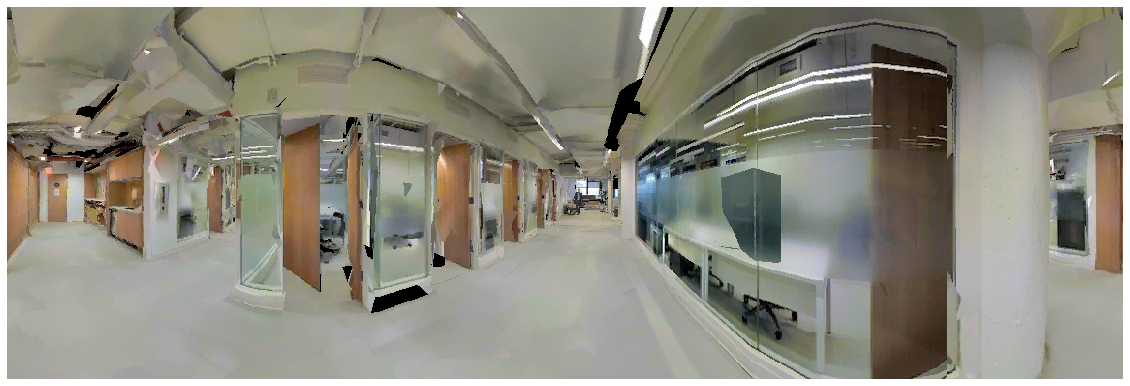}
         \vspace{-6mm}
         \caption{Long Corridor}
         % \label{fig:corridor}
     \end{subfigure}
     \\
     \begin{subfigure}[b]{0.48\textwidth}
         \centering
         \includegraphics[width=\textwidth]{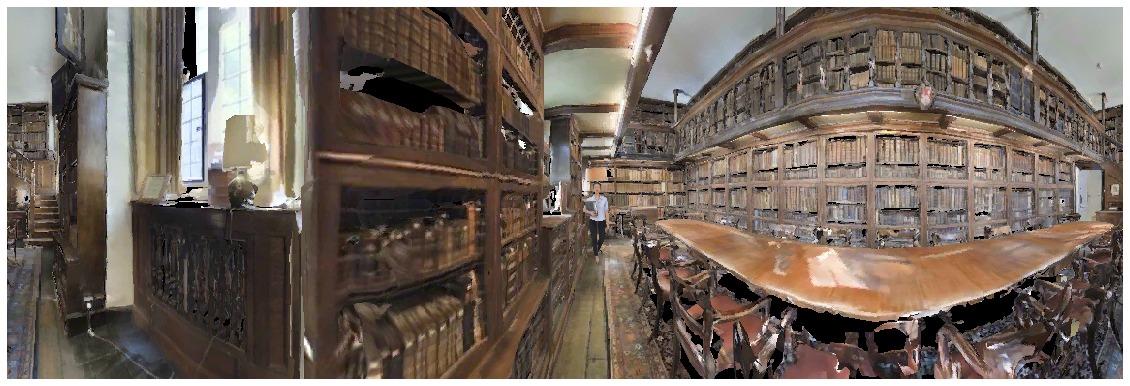}
         \vspace{-6mm}
         \caption{Library}
         % \label{fig:library}
     \end{subfigure}
    \hfill
    \begin{subfigure}[b]{0.48\textwidth}
         \centering
         \includegraphics[width=\textwidth]{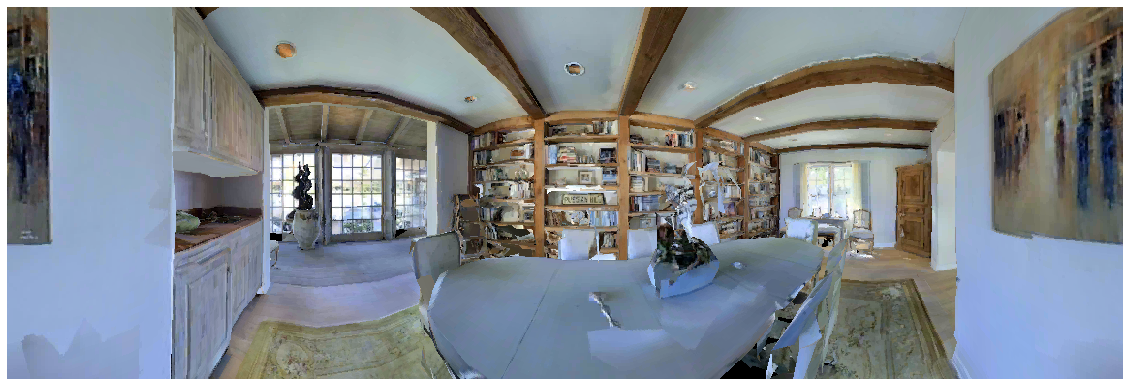}
         \vspace{-6mm}
         \caption{Meeting room}
         % \label{fig:meeting_room}
     \end{subfigure}
        \caption{Some examples from the Matterport 3D dataset showing different challenging environments the model was evaluated against.}
        \label{fig:samples}
\end{figure*}

\subsection{Evaluation Details}
{\noindent \textbf{Task Description.}} As mentioned in Section 5, to compare the performance of AdVerb with the baselines under consideration, we evaluate our models on 3 speech tasks, including automatic Speech Recognition (SR), Speaker Verification (SV), and Speech Enhancement (SE). These 3 tasks can be explained as follows:

\begin{itemize}
    \item The objective of \textbf{SE} is to improve the overall speech quality by suppressing the noise in a noisy speech signal. We evaluate performance on this task using the standard Perceptual Evaluation of Speech Quality (PESQ) metric.
    \item The goal of \textbf{SR} is to automatically transcribe a given speech signal into its corresponding text or contents of the speech utterance. We evaluate performance on this task using the standard Word Error Rate (WER), which calculates the word-level edit distance between the transcribed output and the ground truth text.
    \item Aim of \textbf{SV} is to distinguish if two utterances were from two distinct speakers. We evaluate performance on this task using the standard Equal Error Rate (EER). The EER is the location on a ROC or DET curve where the false acceptance rate and false rejection rate are equal.
\end{itemize}

% \vspace{1mm}
% \noindent \textbf{Hyperparameter Tuning:} 
\subsection{Hyperparameter Tuning Experiments}
We train AdVerb for 100 epochs with a batch size of 16 using an Adam optimizer. For model optimization, we find $\lambda$ = 1 and $\mu$ = 0.1 give the best performance. All hyperparameters were tuned with grid search for the best performance on the dev set. In this sub-section, we show the performance of our model with different  values of $\lambda$ and $\mu$ for Spectrogram Prediction and Acoustic Token Matching losses respectively. Table \ref{tab:lambda} shows the effect of $\lambda$ on AdVerb performance while $\mu$ is kept constant at 0.1. As we clearly see, the performance across all 3 tasks degrades as $\lambda$ decreases. Table \ref{tab:mu} shows the effect of $\mu$ on AdVerb performance while $\lambda$ is kept constant at 0.1. The performance of ASR falls sharply where $\mu$ moves to zero, which proves that the Acoustic Token Matching loss helps preserve phonetic information in speech, thereby improving ASR performance. All experiments were done for the non-fine-tuned version of our experimental setup, where a pre-trained model was used from SpeechBrain.

\begin{table}[h]
\centering
\begin{tabular}{@{}l|cccc}
% \toprule
\toprule
 & \textbf{SE} & \textbf{ASR} & \textbf{SV}     \\
  & \textbf{PESQ ↑} & \textbf{WER ↓} & \textbf{EER ↓} \\
\midrule
$\lambda$ = 1  & 2.96    & 3.54     & 3.11  \\
$\lambda$ = 0.1  & 2.11   & 3.97     & 3.76  \\
$\lambda$ = 0.01  & 2.02   & 4.13    & 4.04  \\
$\lambda$ = 0.001  & 1.99    & 5.01     & 4.11   \\ 
$\lambda$ = 0  &  1.97   & 5.87    & 4.21  \\ \bottomrule
\end{tabular}
\vspace{1mm}
\caption{Result comparison of AdVerb for different values of $\lambda$ across 3 speech tasks on LibriSpeech. All settings are consistent with Section 5 in the paper.}
\label{tab:lambda}
\end{table}

\begin{table}[h]
\centering
\begin{tabular}{@{}l|cccc}
% \toprule
\toprule
 & \textbf{SE} & \textbf{ASR} & \textbf{SV}     \\
  & \textbf{PESQ ↑} & \textbf{WER ↓} & \textbf{EER ↓} \\
\midrule
$\mu$ = 1  & 2.87    & 3.36     & 3.27  \\
$\mu$ = 0.1  & 2.96    & 3.54     & 3.11    \\
$\mu$ = 0.01  & 2.81    & 4.16     & 3.19    \\
$\mu$ = 0.001  & 2.77    & 4.24    & 3.01    \\
$\mu$ = 0 & 2.89    & 4.67     & 3.17   \\ \bottomrule
\end{tabular}
\vspace{1mm}
\caption{Result comparison of AdVerb for different values of $\mu$ across 3 speech tasks on LibriSpeech. All settings are consistent with Section 5 in the paper.}
\label{tab:mu}
\end{table}

\subsection{User Study}

\subsubsection{More Qualitative Results}

Table \ref{tab:user_study2} extends Table \ref{tab:user_study1} to report the subjective evaluation results (30 participants) against the methods in Table \ref{tab:comp_oth}. Note our approach is complementary to the audio-only methods (we use audio-visual inputs) and a direct comparison with these methods might not be fair. However, to study the effectiveness of our methods we add the following comparisons.

\begin{table}[h]
\vspace{-2mm}
\centering
 \setlength{\tabcolsep}{1.4pt}
\scalebox{0.92}{
\begin{tabular}{@{}l|c|l@{}}
% \toprule
\toprule
\textbf{Baseline}   & \textbf{SoundSpaces(in \%)} & \textbf{AVSpeech(in \%)} \\
\textbf{Method}   & \textbf{(A\% / B\% / C\%)}& \textbf{(A\% / B\% / C\%)}\\ \midrule
Audio-only AdVerb & 20.0 / 6.6 / \textbf{\underline{73.3}}                & \hfill 23.3 / 6.6 / \textbf{\underline{70.0}}             \\
DEMUCS \cite{demucs}    &  13.3 / 10.0 / \textbf{\underline{76.6}}              &  \multicolumn{1}{c}{\hfill 16.6 / 10.0 / \textbf{\underline{73.3}}}            \\
VoiceFixer \cite{voicefixer}   & 30.0 / 6.6 / \textbf{\underline{63.3}}               & \hfill 23.3 / 10.0 / \textbf{\underline{66.7}}               \\
HiFi-GAN \cite{h-gan}         & 16.6 / 3.3 / \textbf{\underline{80.0}}                & \hfill 13.3 / 6.7 / \textbf{\underline{80.0}}              \\
Kothapally \textit{et al.} \cite{kothapally2022complex} & 26.6 / 6.6 / \textbf{\underline{66.6}}                & \hfill 26.6 / 13.3 / \textbf{\underline{60.0}}             \\
\bottomrule
\end{tabular}}
% \vspace{2mm}
\caption{User study results. Where (\textit{\textbf{A,B,C}}) consistent with Table \ref{tab:user_study1}. Users find samples from AdVerb to be perceptually better and cleaner when compared against prior methods}
\label{tab:user_study2}
\vspace{-3.2mm}
\end{table}

\begin{figure}
    \centering
    \fbox{\includegraphics[width=7.5cm]{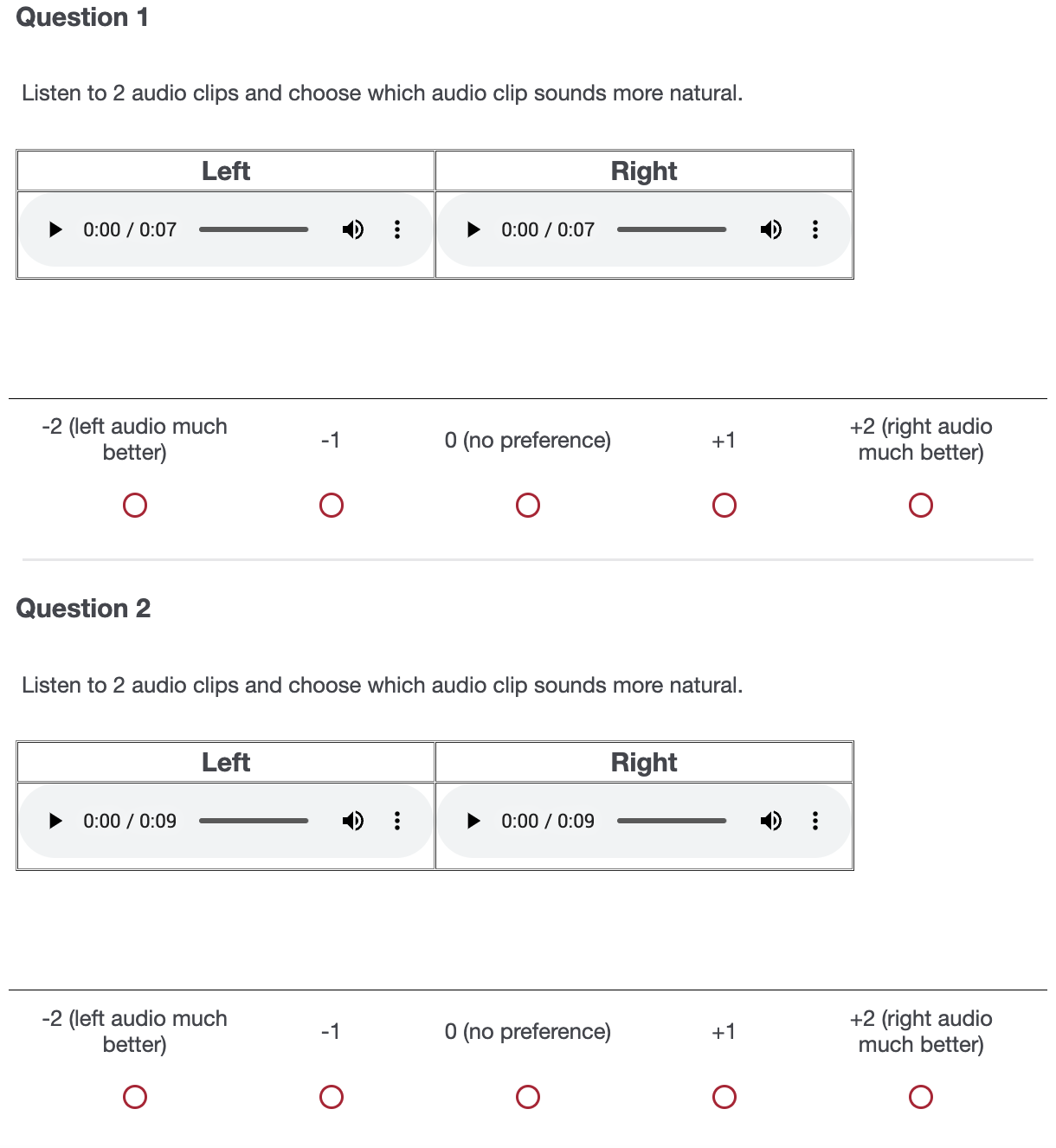}}
    \caption{User study interface. Estimated dereverbed audio samples generated by \modelname~ are compared against that of 3 other baseline models and the GT clean speech (not present for AVSpeech) VIDA\cite{chen2021learning}, WPE\cite{nakatani2010speech} and SCGAN\cite{kothapally2022skipconvgan}. The audio samples are shuffled so that there is no bias among users while rating the audio samples. Participants are asked to choose the audio sample that sounds cleaner and more realistic.}
    \label{fig:survey_diagram}
    \vspace{-5mm}
\end{figure}

{\noindent \textbf{Interface and Evaluation.}} We compare the predicted dereverbed audio produced by \modelname~ with three other SOTA dereverberation models VIDA~\cite{chen2021learning}, SkipConvGAN \cite{kothapally2022skipconvgan}, and WPE \cite{nakatani2010speech}. Fig. \ref{fig:survey_diagram} shows the interface for our user study on Amazon MTurk.

\subsection{Visual Environments and Sample Curation}
Some examples of different environment settings to which our model was evaluated are presented in Fig.\ref{fig:samples}. The dataset proposed in \cite{chen2021learning} uses visual environment samples from Matterport 3D \cite{chang2017matterport3d} with SOTA audio simulations done using SoundSpaces \cite{chensoundspaces} to realistically capture the
environments’ spatial effects on real samples of recorded speech. The dataset enables flexibility over various physical environments, listener/ source positions as well as speech content of the sources. Matterport offers diverse and complex real-world 3D environments with each environment having multiple rooms spanning an average of 517m${^2}$. LibriSpeech \cite{panayotov2015librispeech} which is widely used for benchmarking in the speech recognition literature, was chosen as the source speech corpus. It contains 1,000 hours of 16kHz read English speech from audiobooks. Following \cite{chen2021learning} we train our models with the train-clean-360 split and use the dev-clean and test-clean sets validation and test phases respectively. These splits have non-overlapping speaker identities. Disjoint train/val/test splits for the Matterport 3D visual environments was followed to ensure the house's speaker voices are observed either during training or testing.

\subsection{Societal Impact}

We believe audio-visual dereverberation can positively influence a myriad of real-world applications involving: teleconferencing systems, speech recognition, hearing aids, and video editing among others. Specifically, dereverberation is very critical for hands-free phones and desktop conferencing terminals because, as the microphones are not close to the sound source in these cases but at a considerable distance. 

Lastly, we would like to mention, the user study protocol was approved by Institutional Review Board and we do not collect, share or store any personal information of the participants.

\end{document}